%% file: main.tex
\documentclass[lettersize,journal]{IEEEtran}
\input{main/packages}

\begin{document}
\include{main/commands}
\title{High-resolution open-vocabulary \\ object 6D pose estimation}
\author{Jaime Corsetti, Davide Boscaini, Francesco Giuliari, Changjae Oh, Andrea Cavallaro, and Fabio Poiesi
\thanks{
\textit{(Corresponding author: Jaime Corsetti)}

Jaime Corsetti is with Fondazione Bruno Kessler (FBK), 38123 Trento, Italy, and also with the Department of Information Engineering and Computer Science, University of Trento, 38122 Trento, Italy (e-mail: jcorsetti@fbk.eu).

Davide Boscaini, Francesco Giuliari, and Fabio Poiesi are with Fondazione Bruno Kessler (FBK), 38123 Trento, Italy (e-mail: dboscaini@fbk.eu; fgiuliari@fbk.eu; poiesi@fbk.eu).

Changjae Oh is with the School of Electronic Engineering and Computer Science, Queen Mary University of London, E1 4NS London, United Kingdom (e-mail: c.oh@qmul.ac.uk).

Andrea Cavallaro is with the Idiap Research Institute, CH-1920 Martigny, Switzerland, and also with the École polytechnique fédérale de Lausanne (EPFL), CH-1015 Lausanne,   Switzerland (e-mail: a.cavallaro@idiap.ch).

}}

\maketitle

\input{main/sections/0_abstract}

\input{main/sections/1_intro}
\input{main/sections/2_related}
\input{main/sections/3_method}
\input{main/sections/4_results}
\input{main/sections/5_conclusion}

\bibliographystyle{IEEEtran.bst}
\bibliography{main}
\input{main/sections/6_bios}

\end{document}

%% file: main/packages.tex
\usepackage{amsmath,amsfonts}
\usepackage{algorithmic}
\usepackage{array}
\usepackage[caption=false,font=normalsize,labelfont=sf,textfont=sf]{subfig}
\usepackage{textcomp}
\usepackage{stfloats}
\usepackage{url}
\usepackage{verbatim}
\usepackage{graphicx}
\def\BibTeX{{\rm B\kern-.05em{\sc i\kern-.025em b}\kern-.08em
    T\kern-.1667em\lower.7ex\hbox{E}\kern-.125emX}}
\usepackage{balance}
\usepackage[dvipsnames]{xcolor}
\usepackage{amsmath,amssymb,amsfonts}
\usepackage{multirow,multicol}
\usepackage{array}
\usepackage{xcolor,colortbl}
\usepackage{algorithmic}
\usepackage{graphicx}
\usepackage{textcomp}
\usepackage{xspace}
\usepackage{url}
\usepackage{dsfont}
\usepackage{booktabs}
\usepackage{balance}
\usepackage{soul}
\usepackage{todonotes}
\usepackage{pifont}
\usepackage{colortbl}
\usepackage{overpic}
\usepackage{enumitem}
\usepackage{nccmath}
\usepackage{courier}
\usepackage{array}% http://ctan.org/pkg/array
\usepackage{tabularray}
\usepackage{pifont}% http://ctan.org/pkg/pifont
\usepackage{cite}

%% file: main/commands.tex
\newcommand{\fabio}[1]{\todo[color=purple!20, inline, author=Fabio]{#1}}
\newcommand{\davide}[1]{\todo[color=blue!20, inline, author=Davide]{#1}}
\newcommand{\jaime}[1]{\todo[color=red!20, inline, author=Jaime]{#1}}
\newcommand{\changjae}[1]{\todo[color=olive!20, inline, author=Changjae]{#1}}
\newcommand{\andrea}[1]{\todo[color=cyan!20, inline, author=Andrea]{#1}}
\newcommand{\francesco}[1]{\todo[color=orange!20, inline, author=Francesco]{#1}}
\newcommand{\warning}[1]{\textbf{\textcolor{red!75}{#1}}}

\newcommand{\imgencoder}[0]{$\phi_V$\xspace}
\newcommand{\textencoder}[0]{$\phi_T$\xspace}
\newcommand{\guidance}[0]{$\phi_G$\xspace}
\newcommand{\decoder}[0]{$\phi_D$\xspace}
\newcommand{\fusion}[0]{$\phi_{TV}$\xspace}

\newcommand{\fullpose}[0]{$\mathbf{R}, \mathbf{t}$\xspace}
\newcommand{\rotpose}[0]{$\mathbf{R}$\xspace}
\newcommand{\relpose}[0]{$\mathbf{R}, \mathbf{\tilde{t}}$\xspace}

\newcommand{\learnedprompts}[0]{$\mathbf{D}$\xspace}
\newcommand{\learnedembs}[0]{$\mathbf{E}^D$\xspace}

\newcommand{\clipfeats}[1]{$\mathbf{E}^{#1}$\xspace}
\newcommand{\guidancefeats}[2]{$\mathbf{E}^{#1}_{#2}$\xspace}
\newcommand{\costvolume}[1]{$\mathbf{V}^{#1}$\xspace}
\newcommand{\costemb}[1]{$\mathbf{C}^{#1}$\xspace}
\newcommand{\finalfeats}[1]{$\mathbf{F}^{#1}$\xspace}
\newcommand{\promptemb}[0]{$\mathbf{e}^T$\xspace}

\newcommand{\onedim}[1]{$\in \mathbb{R}^{#1}$\xspace}
\newcommand{\twodim}[2]{$\in \mathbb{R}^{#1\times #2}$\xspace}
\newcommand{\threedim}[3]{$\in \mathbb{R}^{#1\times #2\times #3}$\xspace}

\newcommand{\rgb}[1]{RGB$^{#1}$\xspace}
\newcommand{\depth}[1]{D$^{#1}$\xspace}
\newcommand{\pcd}[1]{P$^{#1}$\xspace}
\newcommand{\predmask}[1]{$\mathbf{M}^{#1}$\xspace}
\newcommand{\query}[0]{$Q$\xspace}
\newcommand{\anchor}[0]{$A$\xspace}
\newcommand{\object}[0]{$O$\xspace}
\newcommand{\prompt}[0]{$T$\xspace}

\newcommand{\lowerbetter}[0]{{\color{black!50}{$\,\downarrow$}}}
\newcommand{\higherbetter}[0]{{\color{black!50}{$\,\uparrow$}}}
\newcommand{\oracle}[1]{\textcolor{gray}{#1}}

\definecolor{visual}{HTML}{3399FF}
\definecolor{text}{HTML}{97D077}

\definecolor{nocsbottle}{RGB}{31,119,180}
\definecolor{nocsbowl}{RGB}{255,127,14}
\definecolor{nocscamera}{RGB}{44,160,44}
\definecolor{nocscan}{RGB}{214,39,40}
\definecolor{nocslaptop}{RGB}{148,103,189}
\definecolor{nocsmug}{RGB}{140,86,75}
\newcommand{\impp}[1]{{\textcolor{Green}{+#1}}}
\newcommand{\impn}[1]{{\textcolor{BrickRed}{-#1}}}

\newcommand{\correct}[1]{{\textcolor{Green}{#1}}}
\newcommand{\wrong}[1]{{\textcolor{BrickRed}{#1}}}

\definecolor{myazure}{rgb}{0.8509,0.8980,0.9412}
\newcommand{\cmark}{\ding{51}}%
\newcommand{\xmark}{\ding{55}}%

\definecolor{mygray}{rgb}{0.9000,0.9000,0.9000}

\newcommand{\captionprompt}[3]{
    \centering \texttt{#1} \\
    \centering \texttt{\correct{#2}} \\
    \centering \texttt{\wrong{#3}}
}

\newcommand{\acronym}{Horyon\xspace}

%% file: main/sections/0_abstract.tex
\begin{abstract}
The generalisation to unseen objects in the 6D pose estimation task is very challenging. 
While Vision-Language Models (VLMs) enable using natural language descriptions to support 6D pose estimation of unseen objects, these solutions underperform compared to model-based methods.
In this work we present \acronym, an open-vocabulary VLM-based architecture that addresses relative pose estimation between two scenes of an unseen object, described by a textual prompt only.
We use the textual prompt to identify the unseen object in the scenes and then obtain high-resolution multi-scale features. 
These features are used to extract cross-scene matches for registration.
We evaluate our model on a benchmark with a large variety of unseen objects across four datasets, namely REAL275, Toyota-Light, Linemod, and YCB-Video.
Our method achieves state-of-the-art performance on {\em all} datasets, outperforming by 12.6 in Average Recall the previous best-performing approach.
\end{abstract}


%% file: main/sections/1_intro.tex
\section{Introduction}\label{sec:intro}

\IEEEPARstart{T}{he} estimation of the 6D pose of an object requires the prediction of its 3D rotation matrix and 3D translation vector with reference to the camera. 
Accurate object 6D pose estimation is a fundamental phase in a wide range of applications, such as augmented reality~\cite{augreality}, robot grasping~\cite{grasping1}, and autonomous driving~\cite{autodriving1}.
Data-driven methods~\cite{fcgf6d,zebrapose} can achieve reliable pose estimation, but require expensive object annotations. 
This problem is mitigated by the development of techniques to generate large-scale synthetic datasets~\cite{he2022fs6d,labbe2022megapose}.
Another line of research instead defines the \textit{unseen-object} setting~\cite{hodan2024bop}, which assumes no overlap between the set of training and test objects~\cite{oneposepp,labbe2022megapose}. 
However, most methods for the unseen-object setting are \textit{model-based}, as they require a CAD model of the object at test time~\cite{zebrapose,pomz,nguyen2023gigapose,osop}, as detailed in the first group of Tab.~\ref{tab:supp_comparison}.
Recent works changed this assumption, and instead require a video sequence of the object at test time~\cite{oneposepp,gen6d}, from which a set of reference views is extracted (second group in Tab.~\ref{tab:supp_comparison}).
These methods use Structure-from-Motion (SfM)~\cite{schoenberger2016sfm} to reconstruct the object 3D model from the reference views~\cite{gen6d,oneposepp}.

To eliminate the need for the object models or multiple views, in our previous work, {Oryon}~\cite{corsetti2023oryon}, we showed that the object can be described with a textual prompt, thus defining the open-vocabulary object 6D pose estimation setting.
This setting assumes the availability of two RGBD scenes, along with a natural language description of the object of interest, provided by the user.
The object 6D pose in a scene is estimated with respect to the object in the other scene.

In this work, we present \acronym (\textbf{H}igh-resolution \textbf{Oryon}), a VLM-based architecture for open-vocabulary object 6D pose estimation that  generalises to unseen objects.
\acronym uses a VLM to extract visual and textual representations from the input scene pair and the natural language description (i.e., the prompt).
The two representations are fused using cross-attention layers that enable information exchange between each scene patch with each prompt token, without losing key details contained in the object description.
The resulting feature maps are upsampled and fused with the high-resolution feature maps from a visual encoder.
\acronym estimates the object pose in the query scene with respect to the anchor scene by segmentation and pixel-level view matching, and subsequently backprojecting and registering the 3D matches.

This work significantly extends our previous work~\cite{corsetti2023oryon} in several aspects.
Instead of processing the whole scene, we identify and localise the object of interest in both scenes with GroundingDino~\cite{liu2023groundingdino} to obtain a bounding box. This leads to a higher resolution  feature map without losing generalisation to unseen objects.
Moreover, we remove the need for the guidance backbone and instead extract the guidance features directly from the VLM, reducing the training time and the number of parameters to 50\% and 37\% of the original method, respectively. 
To validate our choices, we test our method on an extended version of the Oryon benchmark, which includes two popular datasets for pose estimation, i.e., Linemod~\cite{lm} and YCB-Video~\cite{ycbv}, in addition to REAL275~\cite{nocs} and Toyota-Light~\cite{toyl}. 
Linemod features scenes with high clutter, occlusions and objects that are small and unusual.
YCB-Video presents a large variety of poses, and objects that often belong to similar categories (e.g., boxes and cans).
We provide extended ablation studies to justify each choice, and also evaluate the effect of the prompt on the final results.
In summary, our main contributions are:

\begin{itemize}

    \item We provide a comprehensive analysis of recent approaches for estimating the 6D pose of unseen objects, detailing their requirements and operational conditions.

    \item We propose to use a detector that localises and crops the object of interest from the input images using a natural language description. Despite its simplicity, this operation effectively mitigates the influence of extraneous factors, such as background clutter and unrelated objects, on the feature representations used for matching.

    \item We provide a new multi-scale formulation for the VLM-based feature extractor and decoder modules, which enables extraction of higher-quality features for correspondence matching, resulting in an increase of +12.6 in Average Recall compared to the previous version.
    
    \item We extend the evaluation set with two new challenging datasets with occluded (YCB-Video) or small (Linemod) objects. We provide textual prompts for each object and will make this benchmark public.

\end{itemize}

%% file: main/sections/2_related.tex
\section{Related work}\label{sec:related}

\input{main/tables/comparison}

\input{main/figures/diagrams/maindiagram}

In Tab.~\ref{tab:supp_comparison} we report examples of methods for unseen-object 6D pose estimation, along with their assumptions.
We classify them in model-based (first group), model-free (second group), relative pose estimation methods (third group) and open-vocabulary methods (last group).
In the following paragraphs, we describe the assumptions and discuss some example methods for each of the last three groups, as they have the most similar assumptions to \acronym.
We also discuss the usage of localisation modules for pose estimation, to contextualise our choice of an open-vocabulary detector to crop the object.

%%%%%%%%%%%%%%%%%%%%%%%%%%%%%%%%%%%%%%%%%%%%%%%%%%%%%%%%%%%%%%%%%%
\noindent \textbf{Model-free pose estimation.}
Recent research has focused on developing methods for pose estimation of unseen objects without requiring their 3D models during inference, as these 3D models might not always be available.
Model-free methods only require a video sequence captured from multiple viewpoints around the object of interest.
The video sequence is then used to reconstruct an approximate 3D model of the object through structure-from-motion (SfM) techniques~\cite{schoenberger2016sfm}.
Exemplary methods include OnePose++~\cite{oneposepp} and Gen6D~\cite{gen6d}.
OnePose++ employs a graph neural network based on attention~\cite{gatneworks} to establish correspondences between the input image and the 3D reconstruction of the unseen object.
Gen6D leverages three distinct modules for localisation, viewpoint estimation, and pose refinement.
Both methods require a video sequence of the unseen object during inference, assuming the physical availability of the object.
Additionally, a preprocessing procedure is needed to perform the 3D reconstruction of the object, typically involving the execution of an SfM-based algorithm on the video sequence.
This procedure can be cumbersome and challenging for users without technical expertise.
A related approach, FS6D~\cite{he2022fs6d}, instead relies on a sparse set of reference images annotated with their respective camera poses.
In contrast, \acronym does not require annotated images or complex preprocessing procedures; the user only needs to provide a natural language description of the object.

%%%%%%%%%%%%%%%%%%%%%%%%%%%%%%%%%%%%%%%%%%%%%%%%%%%%%%%%%%%%%%%%%%
\noindent \textbf{Relative pose estimation}
approaches estimate the relative pose of an object with respect to one or more reference views~\cite{zhang2022relpose,lin2023relpose++,wang2023posediffusion}.
They are independent on the object's 3D model.
NOPE~\cite{nguyen2023nope} estimates the rotation of an object in a scene given a single reference view, but the usage of RGB information only does not allow it to estimate the translation component of the pose.
Similarly, RelPose~\cite{zhang2022relpose} estimates the relative rotation of an object using as few as three reference views, but it requires that these views come from the same scene. %, with an energy-based formulation.
RelPose++~\cite{lin2023relpose++} enhances its predecessor by introducing a reference frame, enabling the estimation of both the translation and rotation components of the pose using as few as two RGB views of the same scene.
However, RelPose++ only estimates the pose up to a scale factor; the translation component retrieved is scaled under the assumption that the first camera has q unitary distance from the object.
Moreover, RelPose++ requires the images to be approximately centred on the object of interest and does not handle the presence of multiple objects in the same scene.
Similarly to RelPose++, PoseDiffusion~\cite{wang2023posediffusion} addresses camera pose estimation up to a scale factor in the translation component.
It allows for the estimation of camera poses using 2-8 views of the object by enforcing consistency between pairwise matches through geometric constraints.
However, like RelPose++, PoseDiffusion also requires object-centric images and is therefore limited to working with a single object per image.
In contrast, \acronym can estimate an absolute value for the translation component of the pose and does not assume the views are captured from the same scene.

%%%%%%%%%%%%%%%%%%%%%%%%%%%%%%%%%%%%%%%%%%%%%%%%%%%%%%%%%%%%%%%%%%
\noindent \textbf{Language models for pose estimation.} 
So far, the use of VLMs for pose estimation tasks has been limited.
An early approach~\cite{language6dpose} adopted a grounding network to extend the capabilities of a grasping robot in a category-level pose estimation scenario.
Subsequent works similarly used textual prompts for instance-level~\cite{fu2023lanpose} or category-level~\cite{cai2024ov9d} pose estimation, but their generalisation capability is limited to the training categories.
Recently, Oryon~\cite{corsetti2023oryon} has shown an effective approach based on joint image matching and segmentation in order to estimate the pose of a common object given two scenes.
Instead of relying on an object model or a set of reference views, the object of interest is described with a textual prompt, and localised by segmentation. 
The reference scene and the usage of depth information allows Oryon to estimate the relative 6D pose between the two scenes, even for objects unseen at test time. 
However, the limited resolution of Oryon's feature map limits its ability to estimating accurate poses. 
In this work, we address this limitation by proposing a set of key modifications, which include a higher-resolution feature map for fine-grained feature matching, and an updated VLM to better exploit the textual prompt.

%%%%%%%%%%%%%%%%%%%%%%%%%%%%%%%%%%%%%%%%%%%%%%%%%%%%%%%%%%%%%%%%%%
\noindent \textbf{Object localisation for pose estimation.}
Adopting external modules to localise the object of interest within input images is standard practice in the pose estimation community~\cite{ffb6d,zebrapose,nguyen2023gigapose}.
This is often necessary to deal with highly occluded datasets~\cite{lmo}, and is common both in the classic setting~\cite{fcgf6d,zebrapose} and in the unseen-object setting~\cite{onepose,oneposepp}.
Crops are obtained by training a detector on the specific objects (e.g., YOLOv5~\cite{yolov5} in OnePose and OnePose++, FCOS~\cite{tian2020fcos} in ZebraPose~\cite{zebrapose}) or by using ground-truth bounding boxes~\cite{he2022fs6d,lin2023relpose++,wang2023posediffusion}.
Notable exceptions are methods that train the detector contextually to the rest of the pipeline, such as Gen6D~\cite{gen6d} and OSOP~\cite{osop}.
For non-generalisable methods, adopting supervised detectors~\cite{he2017maskrcnn,tian2020fcos,yolov5} to obtain a bounding box is reasonable, as they assume to have access to the object model at test time. 
Therefore using them to train a detector does not lead to loss of generalisation.
Recently, CNOS~\cite{groueix2023cnos} has been proposed to provide accurate segmentation masks given a CAD model of the object, without training on the specific object instances.
This method can be naturally paired with model-based methods for unseen-object pose estimation~\cite{labbe2022megapose,pomz,osop}, as CNOS shares their assumptions (i.e., object 3D model available at test time, but no specific training on it).
\acronym retains the open-vocabulary assumptions, as we use an open-vocabulary detector to locate and crop the object.

%% file: main/tables/comparison.tex
\begin{table*}[t!]
\centering
\tabcolsep 3pt
\caption{
Comparison of the data requirements of \acronym with examples of state-of-the-art methods for unseen-object 6D pose estimation.
We classify the methods based on:
\textbf{Input}: the type of input data, typically RGB or RGBD; 
\textbf{Reference}: additional data used to identify the unseen object at test time;
\textbf{Pose}: whether the method is capable of estimating the 6D pose or is limited to the rotation component. Methods that can estimate the translation component up to a scale are identified by \relpose.;
\textbf{Object preprocessing}: eventual process required at test time to acquire information about the object of interest;
\textbf{Localisation}: eventual external modules used to localise the object, typically a segmentor or a detector;
\textbf{Zero-shot}: True if the localisation module was not specifically trained for the test dataset.
}
\vspace{-3mm}
\label{tab:supp_comparison}
\resizebox{\textwidth}{!}{%
    \begin{tabular}{|l|cc|l|c|l|cc|}
        \toprule
        \multirow{2}{*}{\textsc{\textbf{Method}}} & \multicolumn{2}{c|}{\textsc{\textbf{Input}}} & \multirow{2}{*}{\textsc{\textbf{Reference}}} & \multirow{2}{*}{\textsc{\textbf{Pose}}} & \multirow{2}{*}{\textsc{\textbf{Object preprocessing}}} & \multicolumn{2}{c|}{\textsc{\textbf{Localisation}}} \\
        & \textsc{\textbf{RGB}} & \textsc{\textbf{D}} & & & & \textsc{\textbf{Type}} & \textsc{\textbf{Zero-shot}} \\
        \hline
        OVE6D~\cite{ove6d} & & \cmark & 3D model & \fullpose & Generates and encodes 4K object templates & Segmentor & \xmark \\
        MegaPose~\cite{labbe2022megapose} & \cmark & \cmark & 3D model & \fullpose & - & Detector & \xmark \\
        OSOP~\cite{osop} & \cmark &  & 3D model & \fullpose & Generates 90 object templates & - & - \\
        \hline
        Gen6D~\cite{gen6d} & \cmark &  & Video sequence & \fullpose & SfM and manual cropping of point cloud & - & - \\
        OnePose~\cite{onepose} & \cmark &  & Video sequence & \fullpose & SfM to retrieve camera viewpoints & Detector & \xmark \\
        OnePose++~\cite{oneposepp} & \cmark & & Video sequence & \fullpose & SfM to retrieve camera viewpoints & Detector & \xmark \\
        \hline
        NOPE~\cite{nguyen2023nope} & \cmark & & Single supp. view & \rotpose & - & - & - \\
        RelPose~\cite{zhang2022relpose} & \cmark & & Two or more supp. views & \rotpose & - & - & - \\
        RelPose++~\cite{lin2023relpose++} & \cmark & & One or more supp. views & \relpose & - & Detector & \xmark \\
        PoseDiffusion~\cite{wang2023posediffusion} & \cmark & & One or more supp. views & \relpose & - & Detector & \xmark \\
        LatentFusion~\cite{park2020latentfusion} & \cmark & \cmark & One or more supp. views & \fullpose & - & Segmentor & \xmark \\
        \hline
        Oryon & \cmark & \cmark & Single supp. view & \fullpose & Expression of textual prompt & - & - \\
        \acronym (ours) & \cmark & \cmark & Single supp. view & \fullpose & Expression of textual prompt & Detector & \cmark \\
        \bottomrule
    \end{tabular}
}
\end{table*}

%% file: main/figures/diagrams/maindiagram.tex
\begin{figure*}[t]
    \centering

    \begin{overpic}[clip,trim=1.2cm 1cm 1.2cm 1cm, width=1\linewidth]{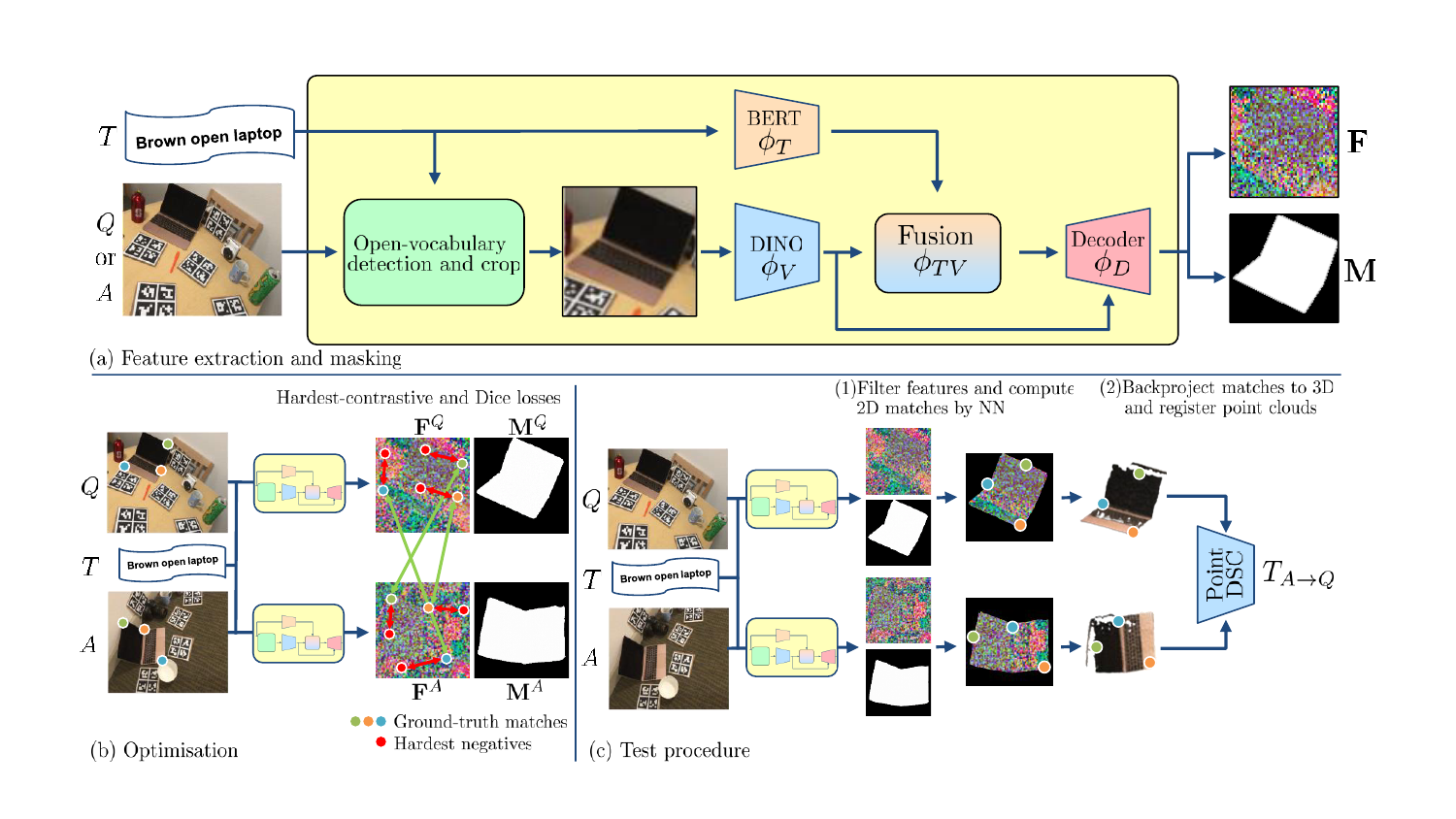}
                
    \end{overpic}

    \caption{
    The main modules of our proposed method, \acronym. (a) Overview of a processing branch:
    \acronym first crops the object of interest from the scene given a textual prompt, and subsequently extracts visual and textual features with DINO and BERT, respectively from the image crop and the prompt.
    The fusion module \fusion outputs a multimodal representation of the scene, which is upsampled by a decoder \decoder.
    At this stage, skip connections from the image encoder \imgencoder are used to enrich the final representation.
    The output features \finalfeats{} are used to obtain the object segmentation mask \predmask{}.
    (b) Optimisation procedure. 
    \finalfeats{A}, \finalfeats{Q} are optimised by a hardest contrastive loss which uses ground-truth matches as supervision, while the segmentation is supervised by a Dice loss.
    (c) Test procedure. 
    The predicted masks are used to filter \finalfeats{A}, \finalfeats{Q}, and matches are obtained by nearest neighbor.
    Finally, the matches are backprojected in 3D, and a registration algorithm is used to retrieve the final pose $T_{A \rightarrow{Q}}$.
    }
    \label{fig:pipeline}
\end{figure*}

%% file: main/sections/3_method.tex
\section{Proposed approach: \acronym}
\label{sec:method}

\begin{figure*}[]
    \centering

    \includegraphics[clip, trim=1.5cm 4.3cm 1.5cm 1cm,width=\linewidth]{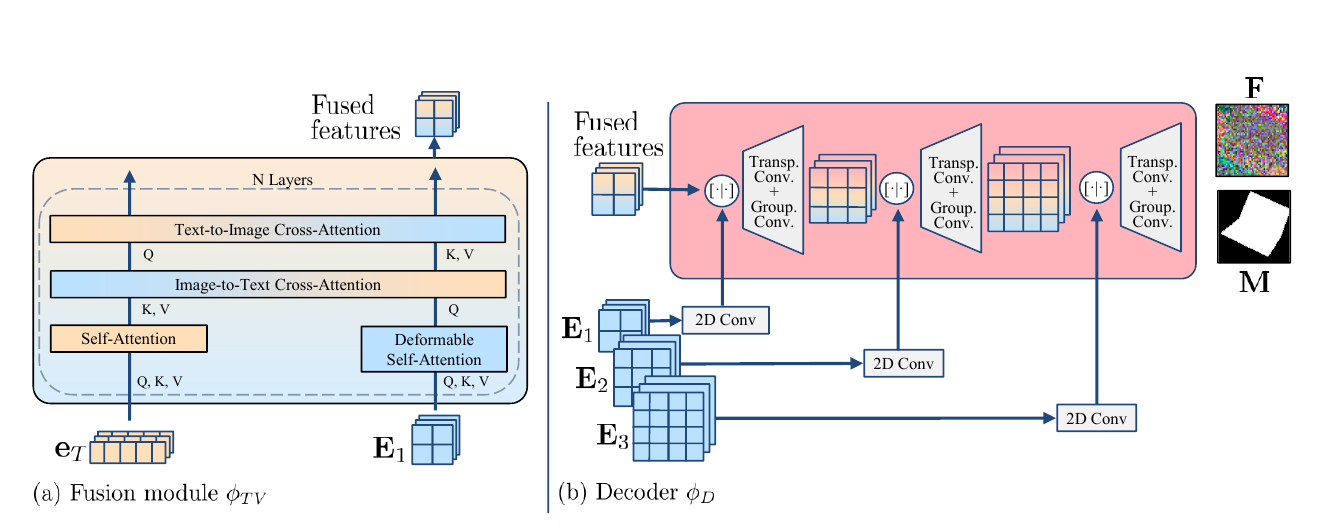}
    \caption{
        (a) Overview of a layer of the fusion module \fusion. 
        Note that the left-side output of the module is not used in the last layer.
        (b) Architecture of the decoder \decoder.
        \guidancefeats{}{1},\guidancefeats{}{2},\guidancefeats{}{3}: visual features obtained from \imgencoder;
        \textbf{Transp.Conv.}: transposed 2D convolution; \textbf{Group.Conv.}: group composed by two blocks, each contains a 2D convolution, a group normalisation and a ReLU; 
        $[ \cdot | \cdot ]$ denotes feature concatenation.
    }
    \label{fig:supp_diagram_decoder}
\end{figure*}

%%%%%%%%%%%%%%%%%%%%%%%%%%%%%%%%%%%%%%%%%%%%%%%%%%%%%%%%%%%%%%%%%%%%%
%%%%%%%%%%%%%%%%%%%%%%%%%%%%%%%%%%%%%%%%%%%%%%%%%%%%%%%%%%%%%%%%%%%%%
\subsection{Overview}
\label{sec:overview}

Let two RGBD scenes, an anchor \anchor and a query \query, depict two scenes with a common object.
Each scene is represented by the RGB channels \rgb{A} (\rgb{Q}) and a depth map \depth{A} (\depth{Q}). 
The intrinsic camera parameters used to capture both \anchor and \query are available.
For each pair (\anchor, \query) a textual prompt \prompt describing the common object \object is provided. 
The object present at test time was not observed at training time.
The objective is to estimate the 6D pose of the object in \query with respect to the reference provided by \anchor.

Fig.~\ref{fig:pipeline}(a) shows the proposed \acronym architecture. 
\acronym consists of four main modules: an open-vocabolary detector, the vision and text encoders, the fusion module, and the decoder.  
The open-vocabulary detector identifies the object \object in both scenes.
The two pre-trained networks: the vision encoder \imgencoder and the text encoder \textencoder, encode the cropped \anchor and \query in multi-scale feature maps and generate an embedding of the prompt \prompt, respectively. 
The fusion module \fusion provides a joint representation across the visual and text modalities.
Finally, the decoder \decoder upsamples the features from the fusion module to provide a set of features that can be used for image matching. 
The final features are processed by a convolutional layer which outputs the localisation masks of the object.
In the following sections, we describe each component, along with the training and test procedures.

\subsection{Object identification and cropping}

We perform 6D pose estimation by image matching, therefore the accuracy of the predicted matches is crucial to the final performance.
To this end, we use feature maps that encode the single object of interest instead of the whole scene.
We process each scene with the open-vocabulary object detector GroundingDino~\cite{liu2023groundingdino}, which allows a good accuracy and can generalise to previously unseen objects.
We use the same prompt \prompt given as input to \acronym, and square and resize each bounding box to avoid deformation.
The predicted bounding boxes are used only at test time, and instead we use the ground-truth ones for training.
In this way, we can obtain accurate detections without losing generalisation capability.

\subsection{Vision-Language backbone}

The features used for matching should be (1) conditioned on the textual prompt \prompt, and (2) robust to unseen objects, for which \acronym was not specifically trained.
We use DINO~\cite{zhang2022dino} as vision encoder \imgencoder to extract multi-scale visual feature maps \guidancefeats{}{1}, \guidancefeats{}{2}, \guidancefeats{}{3} and BERT~\cite{devlin2018bert} as text encoder \textencoder, to encode the prompt in a sequence of textual features \promptemb.
Previous works~\cite{liu2023groundingdino} have proven this combination to be effective in obtaining fine-grained feature maps with good generalisation capabilities.
Using BERT provides an additional advantage: \promptemb is not a global vector, but instead is a sequence of feature vectors, one for each token.
This allows to preserve the information related to the description of the object, and therefore provides a better representation for the natural language description.
Unlike \acronym, Oryon uses CLIP~\cite{clip} as Vision-Language backbone. 
Although CLIP showed impressive capabilities on tasks based on localisation~\cite{zhou2022maskclip}, it was trained for alignment of global embeddings, and therefore was not designed to provide spatially-informed feature maps.

\subsection{Fusion module}

The fusion module \fusion is based on cross-attention between patches of the visual feature maps \guidancefeats{}{1} and tokens of the prompt \promptemb ~\cite{liu2023groundingdino} (see Fig.~\ref{fig:supp_diagram_decoder}(a)).
This strategy allows \acronym to preserve the information contained in each specific token in the prompt, and therefore the model can learn to associate each visual patch with the most suitable component of the prompt (e.g., a token representing ``red'' in the prompt can easily be associated to red patches on the object).
In contrast, Oryon~\cite{corsetti2023oryon} adopts a global representation from CLIP as textual embedding, thus collapsing in a single feature vector all the information contained in the prompt.
We found the new design to be more effective, particularly when the prompt is noisy (see the ablation study on the prompt type in Tab.~\ref{tab:abl_prompts}.

\subsection{Decoder}

The feature map produced by the fusion module \fusion retains the low resolution of the original input features \guidancefeats{}{1}.
In order to obtain high-resolution features necessary for matching, we use a decoder \decoder based on transposed convolutions to upsample the input features.
Prior to each upsample layer, we concatenate the purely visual features obtained from DINO, respectively \guidancefeats{}{1}, \guidancefeats{}{2} and \guidancefeats{}{3}, to enrich the semantic representation of the Vision-Language backbone.
We define such feature map as \textit{guidance features}.
In Oryon~\cite{corsetti2023oryon}, guidance features are extracted from a pretrained Swin transformer~\cite{liu2021swin}.
Instead, we directly use the features from DINO, therefore substantially reducing the number of parameters.
Moreover, the training strategy of DINO encourages generalisation to novel concepts~\cite{zhang2022dino}, and thus it is more effective than Swin in our task.
The output of the decoder is a high-resolution feature map \finalfeats{}.
To obtain the segmentation mask \predmask{}, we process \finalfeats{} with a convolutional layer.

% %%%%%%%%%%%%%%%%%%%%%%%%%%%%%%%%%%%%%%%%%%%%%%%%%%%%%%%%%%%%%%%%%%%%%
\subsection{Optimisation}
\label{sec:optimization}

We train \acronym with an optimisation procedure based on joint image matching and segmentation.
The feature maps \finalfeats{A}, \finalfeats{Q} are optimised with a hardest-contrastive loss $\ell_F$~\cite{FCGF,fcgf6d}, supervised by ground-truth matches between \anchor and \query.
The positive component $\ell_P$ forces the matches to be close in the feature space, while the negative component $\ell_N$ increases the distance between a feature point and its \emph{hardest negative}.

Let $\mathbf{f}^A$, $\mathbf{f}^Q$ \twodim{C}{D} be the features extracted respectively from the feature maps \finalfeats{A}, \finalfeats{Q}, by using the ground-truth matches $\mathcal{P}$.
$C = \lvert \mathcal{P}\rvert$ is the total number of matches and $D$ is the feature dimension. 
The positive component $\ell_P$ is
%+++++++++++++++++++++++++++++++++++++++
\begin{equation}\label{eq:hcpos}
\ell_P = \sum_{(i, j) \in \mathcal{P}} \frac{1}{\lvert \mathcal{P} \rvert} \left( \mathrm{dist} ( \textbf{f}^A_i, \textbf{f}^Q_j) - \mu_P \right)_+,
\end{equation}
%+++++++++++++++++++++++++++++++++++++++
where $( \cdot )_+ = \max(0, \cdot)$ and $\mu_P$ is a positive margin, i.e., the desired distance in the feature space of a positive pair.

We consider a set of features $\mathbf{f}$ and their corresponding coordinates $\textbf{x}$ on the scene to define the negative pairs. 
Given $\mathbf{f}_i$, its candidate negative set is defined as $\mathcal{N}_i = \{ k \colon \textbf{x}_k \in \textbf{x}, k \ne i,  \lVert\textbf{x}_i - \textbf{x}_k\rVert\ \geq \tau\}$.
This excludes all points closer than the distance $\tau$ from the reference point $\textbf{x}_i$ in the scene, thereby excluding features describing the same points.
Candidate negatives sets are computed for all points of $\mathbf{x}^A$ and $\mathbf{x}^Q$, and the negative component $\ell_N$ is
%+++++++++++++++++++++++++++++++++++++++
\begin{equation}\label{eq:hcneg}
\begin{aligned}
\ell_N = \sum_{(i, j) \in \mathcal{P}} & \frac{1}{2 \lvert \mathcal{P}_i \rvert} \left( \mu_N - \min_{k \in \mathcal{N}_i} \mathrm{dist}(\textbf{f}_i, \textbf{f}_k) \right)_+ \\
& \hspace{-4.5mm}+ \frac{1}{2 \lvert \mathcal{P}_j \rvert} \left( \mu_N - \min_{k \in \mathcal{N}_j} \mathrm{dist}(\textbf{f}_j, \textbf{f}_k) \right)_+.
\end{aligned}
\end{equation}
%+++++++++++++++++++++++++++++++++++++++

For each $\mathbf{f}_i$, the nearest $\mathbf{f}_k$ in the feature space (i.e.~the hardest negative) is selected.
Given a negative pair, the negative margin $\mu_N$ is the desired distance of the two points in the feature space.
In Eqs.~\eqref{eq:hcpos} and \eqref{eq:hcneg}, $\mathrm{dist}( \cdot )$ is the inverted and normalised cosine similarity.
We implement the segmentation loss $\ell_M$ as a Dice loss~\cite{sudre2017diceloss}.
The final loss $\ell$ is defined as 
%++++++++++++++++++++++++++++++++++
\begin{equation*}
    \ell = \ell_M + \lambda_N \ell_N + \lambda_{P} \ell_{P},
\end{equation*}
%+++++++++++++++++++++++++++++++++++
\noindent where $\lambda_N$ and $\lambda_P$ are hyperparameters.

In \acronym, we change the loss hyperparameters with respect to the ones used in Oryon, in order to adapt to the new context in which the object is cropped.
Intuitively, a higher resolution implies more ground-truth matches, therefore we raise the number of matches to $C = $ 2000, as opposed to $C = $ 500 in Oryon.
For the same reason, we set the excluding distance for the hardest negatives to $\tau = $ 20.

%%%%%%%%%%%%%%%%%%%%%%%%%%%%%%%%%%%%%%%%%%%%%%%%%%%%%%%%%%%%%%%%%%
\subsection{Matching and registration}

At test time, we obtain the predicted mask \predmask{A}, \predmask{Q} and the features \finalfeats{A}, \finalfeats{Q} from \anchor and \query.
The predicted masks are used to filter the features belonging to the objects, thus obtaining two lists of features $\mathbf{F}^A_M$ \twodim{C^1}{D}, $\mathbf{F}^Q_M$ \twodim{C^2}{D}.

We compute the nearest neighbor $\mathbf{f}^Q_i \in \mathbf{F}^Q_M$ in the feature space for each feature $\mathbf{f}^A_i \in \mathbf{F}^A_M$, and reject the pairs $\mathbf{f}^A_i, \mathbf{f}^Q_i$ for which $\mathrm{dist}(\mathbf{f}^A_i, \mathbf{f}^Q_i) > \mu_T$.

The resulting points are backprojected to the 3D space, by using the depth maps and the intrinsic camera parameters of $A$ and $Q$, thus obtaining two point clouds $\mathbf{P}^A, \mathbf{P}^Q$ \twodim{C}{3}.
Finally,  to obtain the pose $T_{A\rightarrow Q}$, we use PointDSC~\cite{pointdsc} as its spatial consistency formulation allows to reject inconsistent matches, thus leading to more accurate poses.

%% file: main/sections/4_results.tex
\section{Results}\label{sec:exps}

%%%%%%%%%%%%%%%%%%%%%%%%%%%%%%%%%%%%%%%%%%%%%%%%%%%%%%%%%%%%%%%%%%
%%%%%%%%%%%%%%%%%%%%%%%%%%%%%%%%%%%%%%%%%%%%%%%%%%%%%%%%%%%%%%%%%%
\subsection{Experimental setup}
During training, the weights of the image and text encoders \imgencoder, \textencoder are frozen, we only update the weights of the fusion and decoder modules.
We train our model using the Adam optimiser~\cite{adam} for $20$ epochs with learning rate $10^{-4}$, weight decay $5\cdot10^{-4}$, and a cosine annealing scheduler~\cite{cosine} to lower the learning rate to $10^{-5}$.
We randomly augment training data by applying horizontal flipping, vertical flipping, and colour jittering.
The output resolution of \finalfeats{A}, \finalfeats{Q} is $192\times192$, and their feature dimension is $F = 32$.
Loss weights are set as $\lambda_P = \lambda_N = 0.5$. 
We set the positive and negative margins in the loss as $\mu_{P} = 0.2$ and $\mu_{N} = 0.9$, and the excluding distance for hardest negatives as $\tau = 20$.
At test time we set the maximum feature distance to identify a match as $\mu_T = 0.25$.
We limit the number of matches to $C = 2000$.
We implement \acronym with PyTorch Lightning~\cite{lightning}. 
We set the batch size to $8$ and train on four Nvidia V100 GPUs.
In this standard setting, a training requires about $12$ hours.

%%%%%%%%%%%%%%%%%%%%%%%%%%%%%%%%%%%%%%%%%%%%%%%%%%%%%%%%%%%%%%%%%%
%%%%%%%%%%%%%%%%%%%%%%%%%%%%%%%%%%%%%%%%%%%%%%%%%%%%%%%%%%%%%%%%%%
\subsection{Datasets}
We train on ShapeNet6D~\cite{he2022fs6d}, the same synthetic dataset used by Oryon~\cite{corsetti2023oryon}.
For evaluation, we introduce a novel benchmark that extends the one proposed in Oryon.
This comprises four real-world datasets: REAL275~\cite{nocs} and Toyota-Light~\cite{toyl}, which are also part of the original benchmark, as well as two additional datasets, Linemod~\cite{lm} and YCB-Video~\cite{ycbv}, featuring more severe occlusions, clutter and object variety.
To comply with our relative pose estimation setting, we form image pairs (\anchor, \query) by randomly sampling \anchor, \query from their image distribution while ensuring that they are captured from different scenes.
We extract 2K image pairs from each test dataset. 
To accommodate our open-vocabulary setting, we provide natural language descriptions for each object across the four test datasets.
In the following paragraphs, we detail the training and testing datasets.

%%%%%%%%%%%%%%%%%%%%%%%%%%%%%%%%%%%%%%%%%%%%%%%%%%%%%%%%%%%%%%%%%%
\noindent\textbf{ShapeNet6D} (SN6D for short)~\cite{he2022fs6d}
is a large-scale synthetic dataset comprising a diverse collection of scenes generated by rendering ShapeNet~\cite{shapenet2015} objects in various poses against random backgrounds.
SN6D adhere to our zero-shot assumption, as it does not contain any object instances present in the test datasets.
We provide natural language descriptions for each object of SN6D by leveraging ShapeNetSem~\cite{shapenetsem}, a subset of ShapeNet that includes additional metadata associated with the 3D models, to extract both the object name and a set of synonyms.
To generate an object description, we randomly select either the object name or one of the provided synonyms (e.g., possible synonyms for ``television'' are ``tv'', ``telly'', and ``television receiver'').
This augmentation strategy increases the object description variability and reduces overfitting.
By following this procedure, we collect a set of 20K data tuples (\anchor, \query, \prompt) to form our training set.
Note that $T$ contains only semantic information and does not include additional details such as the object colour, material, or physical attributes.

%%%%%%%%%%%%%%%%%%%%%%%%%%%%%%%%%%%%%%%%%%%%%%%%%%%%%%%%%%%%%%%%%%
\noindent\textbf{REAL275}~\cite{nocs}
features 18 objects spanning six different categories, arranged in realistic configurations within diverse indoor scenarios (e.g., on tables, floors).
Its challenges are the presence of multiple instances of the same object categories and a wide variety of viewpoints captured in the scenes.

%%%%%%%%%%%%%%%%%%%%%%%%%%%%%%%%%%%%%%%%%%%%%%%%%%%%%%%%%%%%%%%%%%
\noindent\textbf{Toyota-Light} (TOYL for short)~\cite{toyl}
contains scenes where one of 21 different objects is randomly positioned on various fabric types.
The images are captured under challenging lighting conditions, which is particularly relevant in our setting.
As we process pairs of images, significant lighting differences between them pose a major challenge for image matching.

%%%%%%%%%%%%%%%%%%%%%%%%%%%%%%%%%%%%%%%%%%%%%%%%%%%%%%%%%%%%%%%%%%
\noindent\textbf{Linemod}~\cite{lm} (LM for short)
comprises 15 different objects arranged on a table in various configurations, along with other objects acting as clutter.
It is challenging because the objects are often small, poorly textured with mostly uniform colours, and less conventional compared to those in REAL275 (e.g., plastic toy figures of an ape and a cat).

%%%%%%%%%%%%%%%%%%%%%%%%%%%%%%%%%%%%%%%%%%%%%%%%%%%%%%%%%%%%%%%%%%
\noindent\textbf{YCB-Video}~\cite{ycbv} (YCBV for short)
includes 22 household objects, mainly boxes or cans, arranged in 12 different scenes with distractor objects in the background.
Many objects share similar geometries, making photometric information crucial for accurate identification and pose estimation.
It also contains occlusions, including objects stacked atop one another.

%%%%%%%%%%%%%%%%%%%%%%%%%%%%%%%%%%%%%%%%%%%%%%%%%%%%%%%%%%%%%%%%%%
%%%%%%%%%%%%%%%%%%%%%%%%%%%%%%%%%%%%%%%%%%%%%%%%%%%%%%%%%%%%%%%%%%
\subsection{Evaluation metrics}\label{sec:eval_metrics}
We evaluate the poses using Average Recall~\cite{hodan2024bop} (AR) and ADD(S)-0.1d (abbreviated as ADD)~\cite{hodavn2016evaluation}.
AR measures pose error using three metrics: VSD (Visible Surface Discrepancy), MSSD (Maximum Symmetry-aware Surface Distance), and MSPD (Maximum Symmetry-Aware Projection Distance).
For each metric, AR computes the average recall over a set of thresholds.
For VSD and MSSD, the thresholds range from 5\% to 50\% of the object's diameter, while for MSPD, the thresholds range from 5\% to 50\% of the image size.
The final AR score is the average of these individual recalls.
ADD measures pose error as the average of the pairwise distances between the 3D model points transformed according to the ground-truth and predicted poses.
It then computes the recall of pose errors that are smaller than 10\% of the object's diameter~\cite{hodavn2016evaluation}.
Both AR and ADD are designed to be robust to object symmetries.
Compared to AR, ADD's more restrictive threshold makes it more effective in measuring highly accurate poses.
As the quality of the masks influences the matches, we also evaluate the segmentation by mean Intersection-over-Union (mIoU)~\cite{cho2023catseg,liang2023ovseg} across the image pairs (\anchor, \query).

%%%%%%%%%%%%%%%%%%%%%%%%%%%%%%%%%%%%%%%%%%%%%%%%%%%%%%%%%%%%%%%%%%
%%%%%%%%%%%%%%%%%%%%%%%%%%%%%%%%%%%%%%%%%%%%%%%%%%%%%%%%%%%%%%%%%%
\subsection{Comparison procedure}\label{sec:procedure}

We consider three groups of methods for comparison.
As \textit{crop-free} methods, we report the results of ObjectMatch~\cite{gumeli2023objectmatch}, LatentFusion~\cite{park2020latentfusion} and a pipeline built from SIFT~\cite{lowe1999sift} and PointDSC~\cite{pointdsc}.
Additionally, in this group we report Oryon's~\cite{corsetti2023oryon} results.
As \textit{crop-based} methods, we report \acronym along with the results from ObjectMatch, LatentFusion, and SIFT+PointDSC obtained by using a detector to crop the objects.
We refer as these versions of the baselines as ObjectMatch$^\dagger$, LatentFusion$^\dagger$ and SIFT$^\dagger$ respectively.
Furthermore, as \textit{sparse-view} methods we report results with PoseDiffusion~\cite{wang2023posediffusion} and RelPose++~\cite{lin2023relpose++}.
These are RGB-only methods that require an object detector, but no segmentation mask.
To ensure a fair comparison, our main focus is on crop-based methods that use RGBD data (i.e., the crop-based group), as presented in rows 13-21 of Tab.~\ref{tab:baselines}.

%%%%%%%%%%%%%%%%%%%%%%%%%%%%%%%%%%%%%%%%%%%%%%%%%%%%%%%%%%%%%%%%%%
\noindent \textbf{Oryon}~\cite{corsetti2023oryon} serves as our primary baseline for comparison, being the only existing method for open-vocabulary object 6D pose estimation.
In contrast to approaches that use detectors to crop the input images around the object of interest, Oryon processes entire images without requiring any cropping.
We evaluate Oryon's performance using the official checkpoints made available through its repository.

%%%%%%%%%%%%%%%%%%%%%%%%%%%%%%%%%%%%%%%%%%%%%%%%%%%%%%%%%%%%%%%%%%
\noindent \textbf{ObjectMatch}~\cite{gumeli2023objectmatch} was proposed to tackle point cloud registration of scenes with low overlap.
% This is consistent with our scenario, as we assume that the only overlapping section between two scenes is the one showing the query object.
ObjectMatch is based on SuperGlue~\cite{sarlin2020superglue} to estimate the matches, and on a custom pose estimator.
To compare with it, the mask is used to filter the keypoints obtained by the first step (i.e., keypoint estimation with SuperGlue), and the remaining matches are forwarded to the pose estimator model.
When evaluating with the crop (ObjectMatch$^{\dagger}$), we first crop the object according to the predicted box, and then follow the same procedure as above.
We use ObjectMatch's model trained on ScanNet~\cite{dai2017scannet} from the official repository.
In Oryon~\cite{corsetti2023oryon}, results with ObjectMatch were reported by using the mask to crop the image, and subsequently run the method on the resulting crop.

%%%%%%%%%%%%%%%%%%%%%%%%%%%%%%%%%%%%%%%%%%%%%%%%%%%%%%%%%%%%%%%%%%
\noindent \textbf{SIFT}~\cite{lowe1999sift} is used to extract keypoints and descriptors from each image pair, from which matches are computed through descriptor similarity.
We use the mask to filter the matches, and subsequently backproject them to the 3D space.
The final pose is obtained by registering the points with PointDSC.
When evaluating with the crop (SIFT$^{\dagger}$), we first crop the object of interest according to the predicted bounding box, and then follow the same procedure as above.

%%%%%%%%%%%%%%%%%%%%%%%%%%%%%%%%%%%%%%%%%%%%%%%%%%%%%%%%%%%%%%%%%%
\noindent \textbf{LatentFusion}~\cite{park2020latentfusion} is a method for RGBD-based pose estimation of unseen objects. 
Given a set of RGBD support views with associated masks and poses, LatentFusion first builds a latent representation of the object, by aggregating the features of each view.
Subsequently, given a query view of the same object, the optimal pose is obtained by optimising a differentiable renderer with the latent object representation as input.
To compare with LatentFusion we use \anchor to build the representation, and \query as query view. 
Although LatentFusion has been designed to use 8-16 references, the ablation studies shows that it retains a good performance even with a single reference view, as in our setting.
With the crop (LatentFusion$^{\dagger}$), we fed the cropped images of \anchor and \query.

%%%%%%%%%%%%%%%%%%%%%%%%%%%%%%%%%%%%%%%%%%%%%%%%%%%%%%%%%%%%%%%%%%
For sparse-view methods (rows 1-4) we report results with ground-truth detections and the ones obtained from GroundingDino~\cite{liu2023groundingdino} (GDino).
Crop-free methods (rows 5-12) are reported with the ground-truth mask (Oracle) and the one obtained from by Oryon~\cite{corsetti2023oryon}.
For crop-based methods (rows 13-24), we report results with ground-truth mask (Oracle), the one predicted by CATSeg~\cite{cho2023catseg} and the one predicted by \acronym (Ours).
We use the ground-truth detection when evaluating with the ground-truth mask, and the detection obtained from GDino otherwise.

%%%%%%%%%%%%%%%%%%%%%%%%%%%%%%%%%%%%%%%%%%%%%%%%%%%%%%%%%%%%%%%%%%
%%%%%%%%%%%%%%%%%%%%%%%%%%%%%%%%%%%%%%%%%%%%%%%%%%%%%%%%%%%%%%%%%%
\subsection{Quantitative results}\label{sec:quantitative}

\input{main/tables/baselines}

We report our results in Tab.~\ref{tab:baselines}, along with the ones obtained from the baselines and the ones reported in Oryon~\cite{corsetti2023oryon}.
In rows 1-4 we report the results of sparse-view methods, which only use RGB data. 
Note that in their evaluation procedure, both RelPose++~\cite{lin2023relpose++} and PoseDiffusion~\cite{wang2023posediffusion} use the ground-truth translation component of the pose to obtain a translation in meters.
We observe that PoseDiffusion reaches an overall low performance (8.5 average AR when using GDino as detector), while RelPose++ obtains 21.1 AR with the same detection, which is on par with Oryon (row 12) and also SIFT$^{\dagger}$ (row 18).
PoseDiffusion is based on matches between the two views, while RelPose++ uses an energy-based formulation to recover the relative pose.
In the context of our benchmark, in which images are crowded and distractor objects are present, the matches learned by PoseDiffusion fail at recovering an accurate pose, while RelPose++ can better generalise to our setting.
Nonetheless, both methods score significantly lower then \acronym, which surpasses RelPose++ and PoseDiffusion by 12.3 and 24.9 in average AR, respectively.

Rows 5 to 12 show the results obtain from methods that do not use any detection crop.
In this setting Oryon reaches 20.8 average AR, while SIFT obtains 15.7 average AR, thus showing that SIFT matches can lead to good pose estimation performance when paired with a robust registration method (i.e., PointDSC~\cite{pointdsc}).
On the other hand, LatentFusion and ObjectMatch only obtain 12.1 and 5.3 AR, respectively.
We attribute ObjectMatch's low performance to the domain shift: ObjectMatch was trained for registration of scenes with low overlap, and in our datasets the only overlapping part is the one showing the object of interest.
Moreover, the presence of distractor objects may lead to ambiguities in the matches, which results in an overall low score.
LatentFusion was instead trained on renderings of ShapeNet, and therefore it suffers from the domain shift introduced by our benchmark, which show real images.
Moreover, LatentFusion was not tested with predicted segmentation maps, which by nature may be noisy and led to inaccurate latent representations.

Rows 16 to 21 show results with crop-based baselines, which are our main comparison with \acronym.
For most methods, working on the cropped version of the images is beneficial, even when the resulting matches are filtered by an oracle mask: ObjectMatch improves by 3.9 points on average AR (row 13 vs 5), SIFT by 2.2 points (row 16 vs 7), while LatentFusion loses 0.2 points (row 19 vs 10). 
This is reasonable, as the cropping can be considered a normalising operator on the image scale that reduces the effect of the camera pose on the image (i.e., farther objects are smaller).

\acronym beats SIFT$^{\dagger}$, the next best method, by 11.4 average AR when using our predicted mask (row 24 vs 18), while the gap falls to 8.6 points when the mask predicted by CATSeg is used (rows 23 vs 17).
By comparing the mIoU results on rows 23 and 24, we can observe that CATSeg on average outperforms \acronym, as the first reaches 73.0 mIoU against a 70.7 of the latter.
We can gain more insight on this result by observing the mIoU performances separately on each dataset: while \acronym's and CATSeg's results on TOYL are on par, CATSeg performs slightly better on REAL275 (+3.2 mIoU) and is much more accurate on YCBV (+19.7 mIoU).
On the other hand, on LM Oryon outperforms CATSeg by 13.2 mIoU points.
We observe that REAL275 and YCBV contain many different variations of quite common objects (e.g., cans, cups, laptops), while LM is comprised of many unusually objects (e.g., an office hole puncher, a toy ape).
These results suggest that Oryon is more effective than CATSeg in identifying unusual objects, and therefore exhibit better generalisation capabilities in LM.
On the other hand, CATSeg is more effective with common objects, and possibly can easily disambiguate between similar objects.

In rows 23-24 we can observe how \acronym's pose estimation performance changes when a different mask is used.
The better average AR is obtained with our segmentation mask (33.4 vs 29.4), albeit CATSeg scores on average an higher mIoU (73.0 vs 70.7).
This result is even more evident when considering the results of YCBV in which CATSeg produces better masks than our method, but the AR is still higher when our predicted masks are used (20.6 vs 17.9 in average AR).
As observed in Oryon~\cite{corsetti2023oryon}, an accurate segmentation mask is only important up to a point to determine the quality of pose estimation performance.
In \acronym, our joint optimisation procedure enables to learn masks which are optimised to the matching objective, thus resulting in a better performance than the one obtained with an external mask.

Finally, in row 25 we report the increments of \acronym with respect to Oryon, by comparing the settings with predicted masks and detection boxes (i.e., row 24 vs row 12).
While all the changes are positive, the behaviour is different if each dataset is considered separately.
REAL275 shows the largest improvements in pose estimation, with an AR increment of 25.5.
The increment in ADD(S) is even higher (+27.3), showing that \acronym can produce a much higher ratio of very accurate poses.
On the other hand, TOYL exhibits the smaller improvement, as AR and ADD(S) only improve by 3.0 and 4.2 respectively.
This dataset is less affected by the usage of a detector, as it presents a single object for each scene.
It is also the dataset with the largest change in light and point of view across scenes: this suggests that architectural changes can be made to address this specific setting and provide a larger improvement.
LM and YCBV both exhibit consistent improvements in AR (9.8 and 12.0 respectively), albeit their performance is still low compared the REAL275 and TOYL.
We attribute this fact to the clear higher difficult of these datasets compared to the original ones used in Oryon, which is due to high clutter, object occlusion and object similarity.

%%%%%%%%%%%%%%%%%%%%%%%%%%%%%%%%%%%%%%%%%%%%%%%%%%%%%%%%%%%%%%%%%%
%%%%%%%%%%%%%%%%%%%%%%%%%%%%%%%%%%%%%%%%%%%%%%%%%%%%%%%%%%%%%%%%%%
\subsection{Qualitative results}\label{sec:qualitative}

\begin{figure*}[]
    \centering
    \input{main/figures/qualitatives/pose}
    \vspace{-4mm}
    \caption{
    Sample pose results from REAL275~\cite{nocs} (a), Toyota-Light~\cite{toyl} (b), Linemod~\cite{lm} (c) and YCB-Video~\cite{ycbv} (d).
    All the results use crop from GroundingDino~\cite{liu2023groundingdino} and segmentation mask predicted by \acronym.
    We show the object model coloured by mapping its 3D coordinates to the RGB space.
    Query images are darkened to highlight the object poses.
    }
    \label{fig:poses}
\end{figure*}
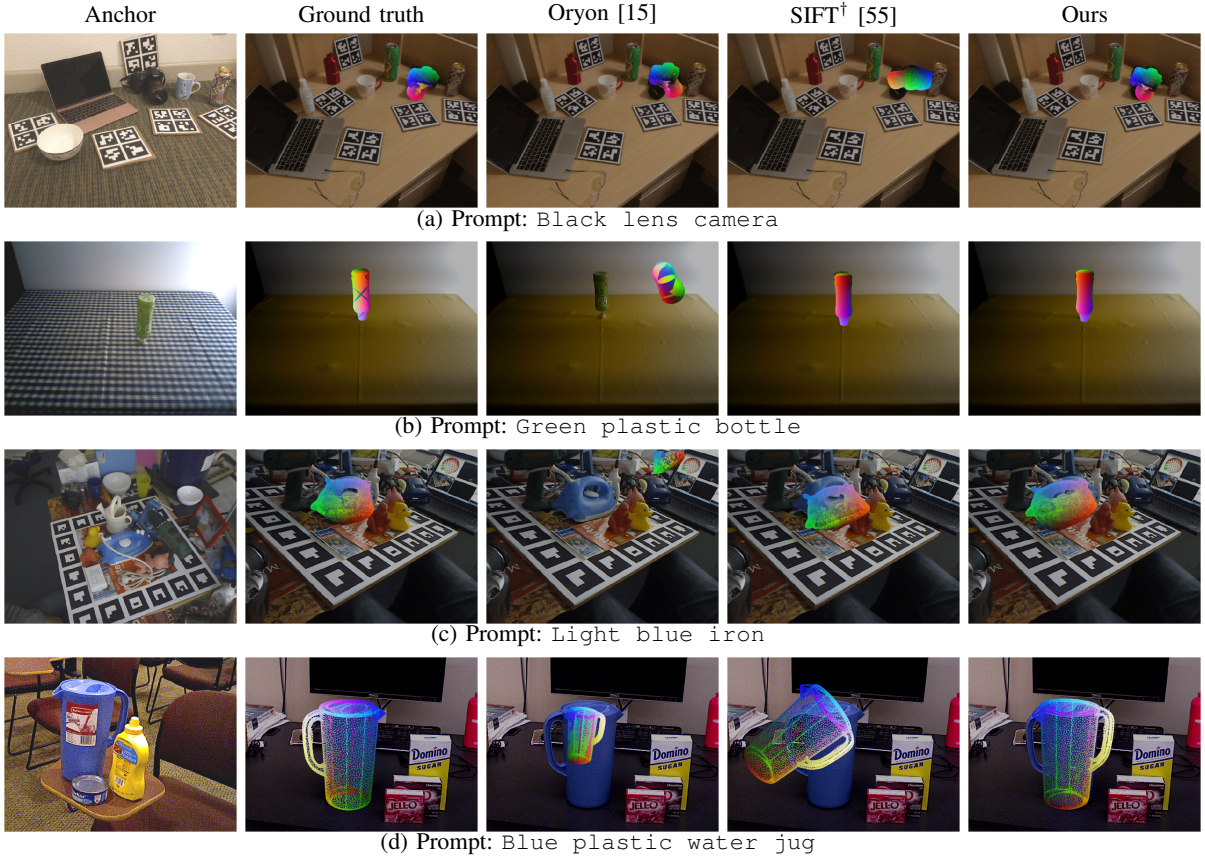

In Fig.~\ref{fig:poses} we report some qualitative result on the pose predicted by \acronym, compared to the ones predicted by Oryon~\cite{corsetti2023oryon}, SIFT$^{\dagger}$~\cite{lowe1999sift} and the ground-truth pose.
Oryon's results are reported using its predicted mask, while results for SIFT$^{\dagger}$ and \acronym are reported using \acronym's predicted mask and detections from GroundingDino.

Fig.~\ref{fig:poses}(a) show results on REAL275~\cite{nocs}.
In this example \acronym obtains fairly accurate poses, albeit the black camera presents a small rotation error, which causes it to be not completely aligned to the ground truth.
SIFT$^{\dagger}$ and Oryon obtain correct localisations, but they present clearly visible rotation and translation errors.%(Fig.~\ref{fig:poses}(a), Oryon).

On Figs.~\ref{fig:poses}(b), we can observe that \acronym is quite accurate on TOYL.
SIFT$^{\dagger}$ similarly obtains good performances, with only small errors in translation.
Instead, Oryon fails completely at localising the object, which results in a large translation error.
We observe that the high variation in lightning conditions makes particularly difficult to obtain correct matches, especially when the images are represented with a low-resolution feature map as in Oryon.

Figs.~\ref{fig:poses}(c) show results on LM.
\acronym retrieves a pose with a small rotation error, which is larger in the prediction of SIFT$^{\dagger}$.
In this case Oryon fails, likely due to a wrong localisation of the blue iron.
This example is quite difficult due to the high variation in viewpoint between the two scenes.

Figs.~\ref{fig:poses}(d) present results on YCBV.
we can observe that Oryon and SIFT$^{\dagger}$ both fail in estimating the pose of the water jug: the first presents a large translation error, while the second shows a wrong rotation.
In this case, \acronym predicts a pose which is mostly aligned, but one of the rotation components is wrong due to the partial symmetry of the object.
This suggest that, while our method benefits from the high-resolution feature map, it still has difficulties in performing matches in small object regions, such as the handle in the water jug.

%%%%%%%%%%%%%%%%%%%%%%%%%%%%%%%%%%%%%%%%%%%%%%%%%%%%%%%%%%%%%%%%%%
%%%%%%%%%%%%%%%%%%%%%%%%%%%%%%%%%%%%%%%%%%%%%%%%%%%%%%%%%%%%%%%%%%
\subsection{Ablation study}\label{sec:ablation}

\input{main/tables/ablation_oryon}

We report in Tab.~\ref{tab:ablation_oryon} an ablation study on the components of \acronym, with the baseline at row 8.

\noindent \textbf{What is the effect of cropping?}
The effect of the crop is strictly related to the change in loss hyperparameters, as using the crop raises the number of matches due to the higher resolution.
Therefore, in rows 1 to 3 we examine how the addition of the crop and the change in hyperparameters influence each other.
In row 1 we do not use any crop and keep the original loss hyperparameters, resulting in average AR similar to Oryon (20.9 vs 20.8) and much worse than our baseline of 33.4 AR.
Only using the crop (row 2) improves the performance, but still results in a significant gap with respect to our baseline (25.3 vs 33.4 AR).
In row 3 we observe that only updating the loss hyperparameters leads to a worse performance than Oryon (17.8 vs 20.8 in AR), thus motivating our choice.
The most significant change in the loss is in the dimension of the excluding kernel of the negative loss, which restricts the pool of negative candidates. 
Without using the crop, a larger kernel is detrimental as it removes a significant portion of the candidate negatives from the object region.

\noindent \textbf{How important is the fusion design?} To study the fusion impact, we remove (row 4) and replace it with the one from Oryon (row 5).
Both experiments result in a worse AR (30.6 and 29.4 vs 33.4 AR), and also the segmentation quality is lower (64.1 and 67.9 vs 70.7 mIoU).
Therefore, a fusion module based on cross-attention on visual and text modalities is more effective than aggregating the cross-modalities similarity map as in Oryon~\cite{corsetti2023oryon}, as shown in row 5.

\noindent \textbf{How important are the decoder design and the guidance features?}
In row 6, we use a different design for the decoder module, which uses two guidance feature maps from DINO instead of three, as the lowest-resolution feature map is discarded.
This results in a drop of 2 AR points with respect to our baseline, which is mostly due to the REAL275 dataset (50.5 vs 57.9 AR).
Intuitively, the low-resolution feature map captures useful information for larger objects, for which the high-resolution could encode noise related to local pixel variations. 
This explains such behaviour in REAL275, which consists on larger objects on average.
In row 7, we remove the skip connections between DINO and the decoder.
While this change does not significantly impact the mask quality (70.6 vs 70.7 mIoU), there is an important drop in pose quality, as the AR drops from 33.4 to 30.6.
Similarly, switching to a Swin Transformer as guidance backbone (row 9) lowers the AR to 31.3.
These results underline the importance of high-resolution features representative of appearance, to counterbalance the semantic content of the embeddings from the VLM.

\noindent \textbf{What is the influence of the VLM choice?}
In rows 10 and 11 we replace our VLM (GDino) with CLIP~\cite{clip} and ALIGN~\cite{jia2021align}, respectively.
Both backbones similarly underperform our default choice, as they score 23.1 and 23.4 AR (CLIP and ALIGN respectively) against 31.3 AR of row 9.
Note that CLIP and ALIGN encode the prompt in a single global embedding, while we use BERT, which outputs a sequence of token-wise features.
This allows \acronym to retain the information related to the object description, which instead may be lost in a global representation, and therefore is beneficial to the type of prompts we use.

\input{main/tables/ablation_prompt}

\noindent \textbf{What is the most important part of the prompt?}
In Tab.~\ref{tab:abl_prompts} we answer this question by changing the type of prompts used at test time.
The experiments in this table only affect the evaluation procedure, as all the training prompts and parameters remains the same.
We evaluate three alternative prompt types: \textit{No name}, in which the object name is replaced by ``object'' in the prompt (e.g., ``brown open object'' instead of ``brown open laptop''); \textit{Misleading}, in which the object description is changed to be different from the object appearance (e.g., ``white closed laptop'' for a laptop that appears brown and open); \textit{Generic}, that only includes the object name, without any description (e.g., ``laptop'').
In the same table we report our baseline results, with the \textit{standard} prompts.
For a fair evaluation, when evaluating with GDino as detector, we use the same alternative prompt type we fed to \acronym.

In row 2, we report the results with the \textit{No name} prompt.
This experiment results in a very significant drop in both AR (-14.7) and mIoU (-18.1) compared with the baseline at row 8.
While the drop is present and significant in all datasets, it is less catastrophic on TOYL.
This dataset is the only one with a single object for each scene, and therefore changing the prompt introduces less ambiguities.
This setting reflects the case in which the user providing the prompt is faced with an unknown object they cannot name, thus resulting in a partial description about the object characteristics.
Row 4 shows that providing a wrong description greatly impacts the average performance, resulting in a drop of -10.9 AR and -14.9 mIoU.
Similarly to the previous prompt, this change has less impact on the TOYL dataset, while LM is more significantly affected, losing 14.7 and 47.4 points in mIoU respectively.
It is clear that in this case the main source of error is due to wrong localisation (either from GDino or from the segmentation mask).
In row 6 the object description is dropped, resulting in the best average results among the ones with alternative prompts, with an AR of 26.1 and an mIoU of 62.0.
While the average drop is still significant compared to the baseline, TOYL is again a notable exception. 
On this dataset, using a generic description is beneficial, as the experiment outperforms our baseline by 3.1 and 3.2 points in AR and mIoU, respectively.
In a context where no ambiguity is possible (i.e., a single object is present), adding a description is detrimental to the pose estimation performance.
Finally, rows 1, 3 and 5 report the results with the ground-truth localisation and segmentation.
We observe that on average the performances are very close to the baseline with the same mask and detector on row 7.
Unsurprisingly, with a perfect localisation \acronym's features are enough to obtain a good performance in pose estimation, even with a suboptimal prompt.

In conclusion, in our architecture the object name is the most important part of the prompt, as only retaining it still provides a good performance, while completely removing it leads to failure, particularly in case of complex scenes with multiple objects.
This experiment highlights an important limitation to the usage of VLMs for pose estimation: such models are still unable to identify and reason about objects given only a description of their visual characteristics.

%% file: main/tables/baselines.tex
\newcolumntype{g}{>{\columncolor{mygray}}c}
\begin{table*}[t!]
    \centering
    \tabcolsep 3pt
    \caption{
    We compare the results of \acronym with our baselines.
    Experiments are reported with different crop and segmentation priors.
    Results in \textbf{bold} and \underline{underlined} represent best and second-best methods when using predicted priors, respectively.
    In the last row we report the increment with respect to Oryon~\cite{corsetti2023oryon}.
    All metrics are the higher the better.
    }
    \vspace{-3mm}
    \label{tab:baselines}
    \resizebox{\textwidth}{!}{%
    \begin{tabular}{crlccgggcccgggcccggg}
    \toprule
    & & \multirow{2}{*}{Method} & \multirow{2}{*}{Crop} & \multirow{2}{*}{Mask} & \multicolumn{3}{c}{\cellcolor{mygray}REAL275} & \multicolumn{3}{c}{Toyota-Light} & \multicolumn{3}{c}{\cellcolor{mygray}Linemod} & \multicolumn{3}{c}{YCB-Video} & \multicolumn{3}{c}{\cellcolor{mygray}Average} \\
    & & & & & AR & ADD & mIoU & AR & ADD & mIoU & AR & ADD & mIoU & AR & ADD & mIoU & AR & ADD & mIoU \\
    \toprule

    %%%%%%%%%%%%%%%%%%%%%%%%%%%%%
    \color{gray} \multirow{4}{*}{\rotatebox{90}{Sparse-view}} & \small{\color{gray} \scriptsize 1}  & \multirow{2}{*}{PoseDiffusion~\cite{wang2023posediffusion}} & \oracle{Oracle} & - & \oracle{9.2} & \oracle{0.8} & - & \oracle{7.8} & \oracle{1.2} & - & \oracle{10.8} & \oracle{1.4} & - & \oracle{7.5} & \oracle{0.8} & - & \oracle{8.8} & \oracle{1.1} & - \\
    & \small{\color{gray} \scriptsize 2} & & GDino & - & 9.5 & 0.8 & - & 8.1 & 1.6 & - & 7.7 & 1.0 & - & 8.5 & 1.1 & - & 8.5 & 1.1 & - \\

    \cmidrule{2-20}
    & \small{\color{gray} \scriptsize 3}  & \multirow{2}{*}{RelPose++~\cite{lin2023relpose++}}  & \oracle{Oracle} & - & \oracle{22.8} & \oracle{11.9} & - & \oracle{30.9} & \oracle{11.6} & - & \oracle{14.6} & \oracle{9.4} & - & \oracle{15.1} & \oracle{10.8} & - & \oracle{20.8} & \oracle{10.9} & - \\
    & \small{\color{gray} \scriptsize 4} & & GDino & - & 23.1 & 12.8 & - & 30.5 & 11.6 & - & 15.1 & 10.8 & - & 15.5 & 10.6 & - & 21.1 & 11.5 & - \\ 

    %%%%%%%%%%%%%%%%%%%%%%%%%%%%%
    \midrule
    \color{gray} \multirow{10}{*}{\rotatebox{90}{Crop-free}} & \small{\color{gray} \scriptsize 5} & \multirow{2}{*}{ObjectMatch~\cite{gumeli2023objectmatch}} & - & \oracle{Oracle} & \oracle{9.7}& \oracle{4.5}& \oracle{100.0}& \oracle{3.5}& \oracle{1.6}& \oracle{100.0}& \oracle{14.2}& \oracle{13.8}& \oracle{100.0}& \oracle{3.7}& \oracle{2.8}& \oracle{100.0}& \oracle{7.8}& \oracle{5.7}& \oracle{100.0} \\
    & \small{\color{gray} \scriptsize 6} & & - & Oryon & 9.2 & 4.2 & 66.5 & 2.9 & 1.1 & 68.1 & 7.5 & 6.8 & 30.1 & 1.6 & 1.3 & 39.2 & 5.3 & 3.4 & 51.0 \\

    \cmidrule{2-20}
    & \small{\color{gray} \scriptsize 7} & \multirow{2}{*}{SIFT~\cite{lowe1999sift}} & - & \oracle{Oracle} & \oracle{34.1}& \oracle{16.4}& \oracle{100.0}& \oracle{30.3}& \oracle{14.1}& \oracle{100.0}& \oracle{17.6}& \oracle{10.0}& \oracle{100.0}& \oracle{18.5}& \oracle{12.8}& \oracle{100.0}& \oracle{25.1}& \oracle{13.3}& \oracle{100.0}\\ 
    & \small{\color{gray} \scriptsize 8} & & - & Oryon & 24.4 & 12.8 & 66.5 & 27.2 & 9.9 & 68.1 & 6.4 & 2.6 & 30.1 & 4.7 & 2.1 & 39.2 & 15.7 & 6.9 & 51.0 \\

    \cmidrule{2-20}
    & \small{\color{gray} \scriptsize 9} & \multirow{2}{*}{LatentFusion~\cite{park2020latentfusion}} & - & \oracle{Oracle} & \oracle{22.9}& \oracle{9.8}& \oracle{100.0}& \oracle{28.2}& \oracle{10.2}& \oracle{100.0}& \oracle{14.5}& \oracle{7.1}& \oracle{100.0}& \oracle{18.0}& \oracle{10.2}& \oracle{100.0}& \oracle{20.9}& \oracle{9.3}& \oracle{100.0}\\
    & \small{\color{gray} \scriptsize 10} & & - & Oryon & 13.7 & 3.8 & 66.5 & 23.1 & 5.0 & 68.1 & 4.8 & 1.0 & 30.1 & 6.7 & 3.5 & 39.2 & 12.1 & 3.3 & 51.0 \\

    \cmidrule{2-20}
    & \small{\color{gray} \scriptsize 11} & \multirow{2}{*}{Oryon~\cite{corsetti2023oryon}} & - & \oracle{Oracle} & \oracle{46.5} & \oracle{34.9} & \oracle{100.0} & \oracle{34.1} & \oracle{22.9} & \oracle{100.0} & \oracle{25.3} & \oracle{20.4} & \oracle{100.0} & \oracle{19.4} & \oracle{12.8} & \oracle{100.0} & \oracle{31.3} & \oracle{22.7} & \oracle{100.0} \\
    & \small{\color{gray} \scriptsize 12} & & - & Oryon & 32.2 & 24.3 & 66.5 & 30.3 & 20.9 & 68.1 & 12.2 & 10.2 & 30.1 & 8.6 & 1.6 & 39.2 & 20.8 & 14.3 & 51.0 \\

    %%%%%%%%%%%%%%%%%%%%%%%%%%%%%
    \midrule
    \color{gray} \multirow{14}{*}{\rotatebox{90}{Crop-based}} & \small{\color{gray} \scriptsize 13} & \multirow{3}{*}{ObjectMatch$^\dag$~\cite{gumeli2023objectmatch}} & \oracle{Oracle} & \oracle{Oracle} & \oracle{20.5}& \oracle{11.1}& \oracle{100.0}& \oracle{8.0}& \oracle{4.1}& \oracle{100.0}& \oracle{12.2}& \oracle{11.1}& \oracle{100.0}& \oracle{6.0}& \oracle{3.7}& \oracle{100.0}& \oracle{11.7}& \oracle{7.5}& \oracle{100.0} \\
    & \small{\color{gray} \scriptsize 14} & & GDino& CATSeg & 21.5 & 11.6 & 84.5 & 7.3 & 4.0 & 81.6 & 13.8 & 11.7 & 54.4 & 6.0 & 3.4 & 71.7 & 12.2 & 7.7 & 73.0 \\
    & \small{\color{gray} \scriptsize 15} & & GDino& Ours& 21.0 & 11.0 & 81.3 & 8.2 & 4.3 & 82.1 & 13.6 & 11.8 & 67.6 & 5.9 & 3.4 & 52.0 & 12.2 & 7.6 & 70.7 \\

    \cmidrule{2-20}
    & \small{\color{gray} \scriptsize 16} & \multirow{3}{*}{SIFT$^\dag$~\cite{lowe1999sift}} & \oracle{Oracle} & \oracle{Oracle} & \oracle{38.8}& \oracle{21.6}& \oracle{100.0}& \oracle{32.4}& \oracle{16.5}& \oracle{100.0}& \oracle{18.7}& \oracle{10.8}& \oracle{100.0}& \oracle{19.3}& \oracle{13.9}& \oracle{100.0}& \oracle{27.3}& \oracle{15.7}& \oracle{100.0} \\
    & \small{\color{gray} \scriptsize 17} & & GDino& CATSeg & 32.9 & 15.2 & 84.5 & 29.5 & 14.8 & 81.6 & 9.4 & 5.0 & 54.4 & 11.3 & 6.1 & 71.7 & 20.8 & 10.3 & 73.0 \\
    & \small{\color{gray} \scriptsize 18} & & GDino& Ours& 33.5 & 18.1 & 81.3 & 29.9 & 14.6 & 82.1 & 14.6 & 8.5 & 67.6 & 10.1 & 6.5 & 52.0 & 22.0 & 11.9 & 70.7 \\

    \cmidrule{2-20}
    & \small{\color{gray} \scriptsize 19} & \multirow{3}{*}{LatentFusion$^\dag$~\cite{park2020latentfusion}} & \oracle{Oracle} & \oracle{Oracle} & \oracle{22.6}& \oracle{9.6}& \oracle{100.0}& \oracle{28.2}& \oracle{10.2}& \oracle{100.0}& \oracle{14.5}& \oracle{6.4}& \oracle{100.0}& \oracle{17.5}& \oracle{10.2}& \oracle{100.0}& \oracle{20.7}& \oracle{9.1}& \oracle{100.0} \\
    & \small{\color{gray} \scriptsize 20} & & GDino& CATSeg & 18.6 & 6.0 & 84.5 & 26.6 & 9.8 & 81.6 & 9.3 & 4.8 & 54.4 & 9.3 & 5.1 & 71.7 & 15.9 & 6.4 & 73.0 \\
    & \small{\color{gray} \scriptsize 21} & & GDino& Ours& 19.8 & 8.2 & 81.3 & 26.0 & 10.3 & 82.1 & 10.3 & 3.9 & 67.6 & 11.0 & 8.9 & 52.0 & 16.8 & 7.8 & 70.7 \\

    \cmidrule{2-20}
    & \small{\color{gray} \scriptsize 22} & \multirow{3}{*}{\acronym} & \oracle{Oracle} & \oracle{Oracle} &\oracle{57.7}& \oracle{49.8}& \oracle{100.0}& \oracle{37.4}& \oracle{28.4}& \oracle{100.0}&\oracle{34.4}& \oracle{27.6}& \oracle{100.0}& \oracle{28.6}& \oracle{22.6}& \oracle{100.0}&\oracle{39.5}& \oracle{32.1}& \oracle{100.0} \\
    & \small{\color{gray} \scriptsize 23} & & GDino& CATSeg & \underline{47.4} & \underline{39.6} & \textbf{84.5} & \underline{32.7} & \underline{23.5} & \underline{81.6} & \underline{19.6} & \underline{17.4} & \underline{54.4} & \underline{17.9} & \underline{9.3} & \textbf{71.7}  & \underline{29.4} & \underline{22.5} & \textbf{73.0} \\
    & \small{\color{gray} \scriptsize 24} & & GDino& Ours & \textbf{57.9} & \textbf{51.6} & \underline{81.3} & \textbf{33.0} & \textbf{25.1} & \textbf{82.1} & \textbf{22.0} & \textbf{20.4} & \textbf{67.6} & \textbf{20.6} & \textbf{12.3} & \underline{52.0} & \textbf{33.4} & \textbf{27.3} & \underline{70.7} \\

    %%%%%%%%%%%%%%%%%%%%%%%%%%%%%
    \midrule
    & \small{\color{gray} \scriptsize 25} & $\Delta$ score & & & \impp{25.5} & \impp{27.3} & \impp{14.8} & \impp{3.0} & \impp{4.2} & \impp{14.0} & \impp{9.8} & \impp{10.2} & \impp{37.5} &\impp{12.0} & \impp{10.7} & \impp{12.8} & \impp{12.6} & \impp{13.0} & \impp{19.7} \\
    \bottomrule        
\end{tabular}  
}
\end{table*} 

%% file: main/figures/qualitatives/pose.tex
% success
\hspace*{0.00mm}
\begin{minipage}{0.17\textwidth}
    \centering{\small{Anchor}}
\end{minipage}
\begin{minipage}{0.17\textwidth}
    \centering{\small{Ground truth}}
\end{minipage}
\begin{minipage}{0.17\textwidth}
    \centering{\small{Oryon~\cite{corsetti2023oryon}}}
\end{minipage}
\begin{minipage}{0.17\textwidth}
    \centering{\small{SIFT$^\dagger$~\cite{lowe1999sift}}}
\end{minipage}
\begin{minipage}{0.17\textwidth}
    \centering{\small{Ours}}
\end{minipage}

% 2 4 1 369 air xin laptop

\vspace*{1.00mm}
%\hspace*{0.00mm}
% % 1 87 5 213 chinese mug
%\vspace*{1mm}
\hspace*{0.00mm}
\begin{minipage}{0.17\textwidth}
    \begin{overpic}[width=1\textwidth]{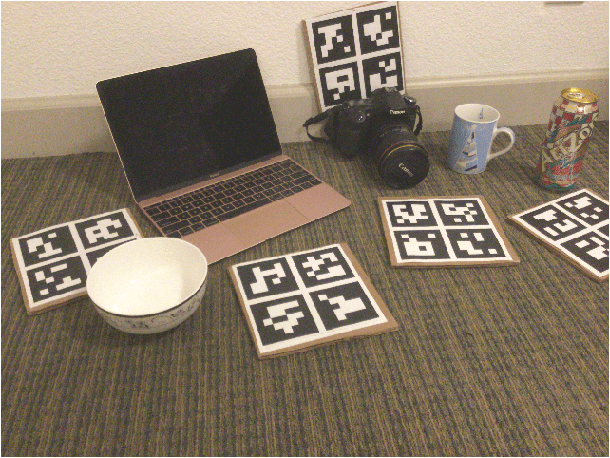}
    % \put(6,43){\small{\textcolor{magenta}{$\mathbf{Q}$}}}
    % \put(54,43){\small{\textcolor{magenta}{$\mathbf{A}$}}}
    \end{overpic}
\end{minipage}
\begin{minipage}{0.17\textwidth}
    \begin{overpic}[width=1\textwidth]{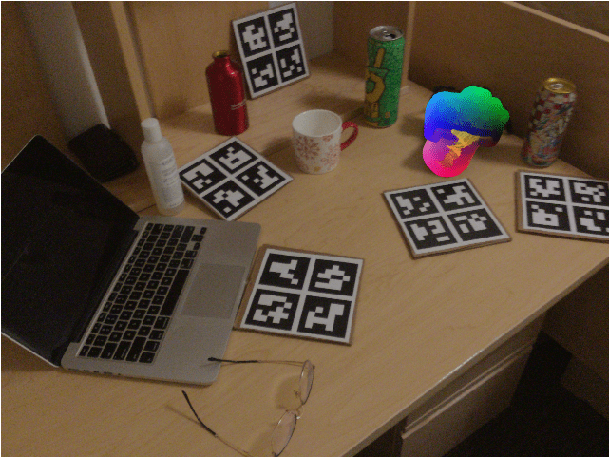}
    % \put(6,43){\small{\textcolor{magenta}{$\mathbf{Q}$}}}
    % \put(54,43){\small{\textcolor{magenta}{$\mathbf{A}$}}}
    \end{overpic}
\end{minipage}
\begin{minipage}{0.17\textwidth}
    \begin{overpic}[width=1\textwidth]{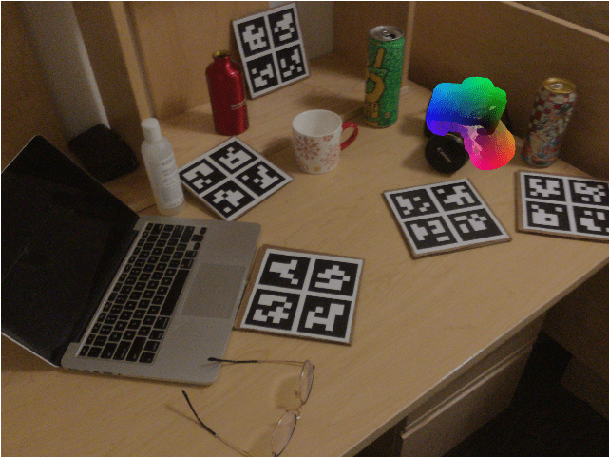}
    % \put(6,43){\small{\textcolor{magenta}{$\mathbf{Q}$}}}
    % \put(54,43){\small{\textcolor{magenta}{$\mathbf{A}$}}}
    \end{overpic}
\end{minipage}
\begin{minipage}{0.17\textwidth}
    \begin{overpic}[width=1\textwidth]{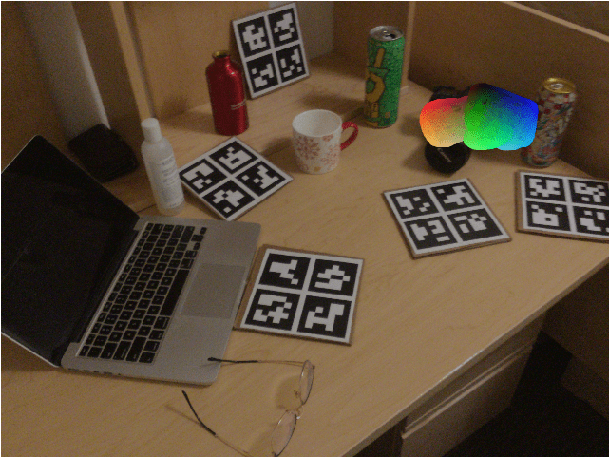}
    % \put(6,43){\small{\textcolor{magenta}{$\mathbf{Q}$}}}
    % \put(54,43){\small{\textcolor{magenta}{$\mathbf{A}$}}}
    \end{overpic}
\end{minipage}
\vspace*{0.2mm}
\begin{minipage}{0.17\textwidth}
    \begin{overpic}[width=1\textwidth]{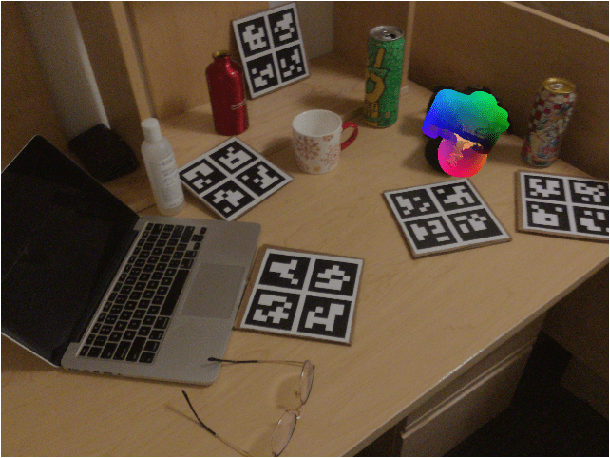}
    % \put(6,43){\small{\textcolor{magenta}{$\mathbf{Q}$}}}
    % \put(54,43){\small{\textcolor{magenta}{$\mathbf{A}$}}}
    \end{overpic}
\end{minipage}
\vspace*{1.5mm}
\begin{minipage}{\textwidth}
    \small \centering{(a) Prompt: \texttt{Black lens camera}}
\end{minipage}

% 3 120 5 198 canon wu camera
% 6 473 3 434 can lotte milk
\hspace*{0.00mm}
\begin{minipage}{0.17\textwidth}
    \begin{overpic}[width=1\textwidth]{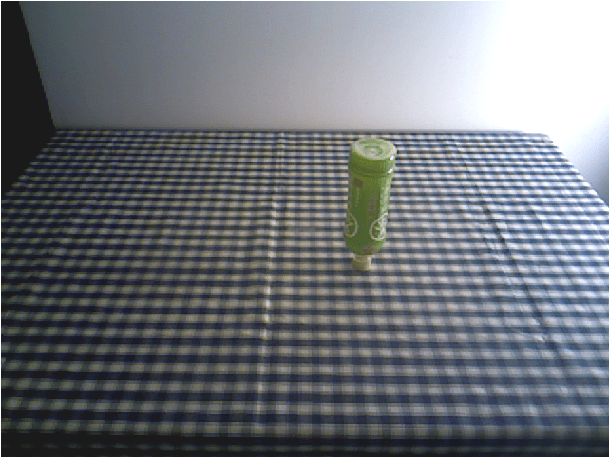}
    % \put(6,43){\small{\textcolor{magenta}{$\mathbf{Q}$}}}
    % \put(54,43){\small{\textcolor{magenta}{$\mathbf{A}$}}}
    \end{overpic}
\end{minipage}
\begin{minipage}{0.17\textwidth}
    \begin{overpic}[width=1\textwidth]{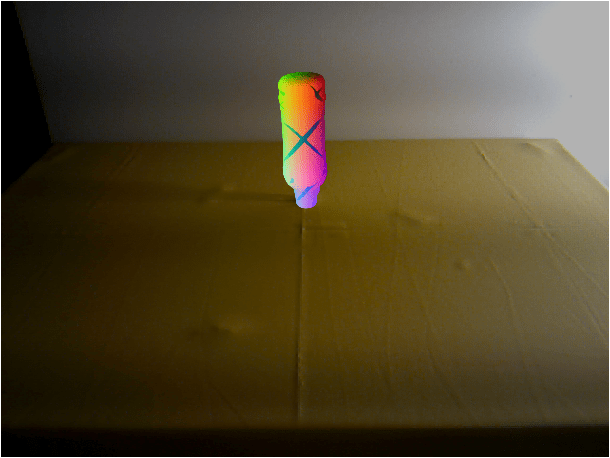}
    % \put(6,43){\small{\textcolor{magenta}{$\mathbf{Q}$}}}
    % \put(54,43){\small{\textcolor{magenta}{$\mathbf{A}$}}}
    \end{overpic}
\end{minipage}
\begin{minipage}{0.17\textwidth}
    \begin{overpic}[width=1\textwidth]{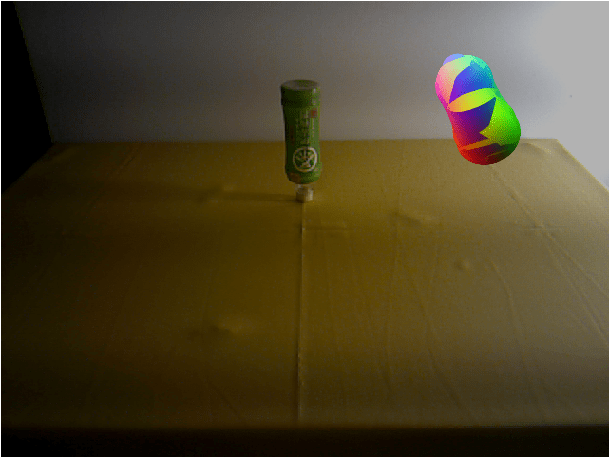}
    % \put(6,43){\small{\textcolor{magenta}{$\mathbf{Q}$}}}
    % \put(54,43){\small{\textcolor{magenta}{$\mathbf{A}$}}}
    \end{overpic}
\end{minipage}
\begin{minipage}{0.17\textwidth}
    \begin{overpic}[width=1\textwidth]{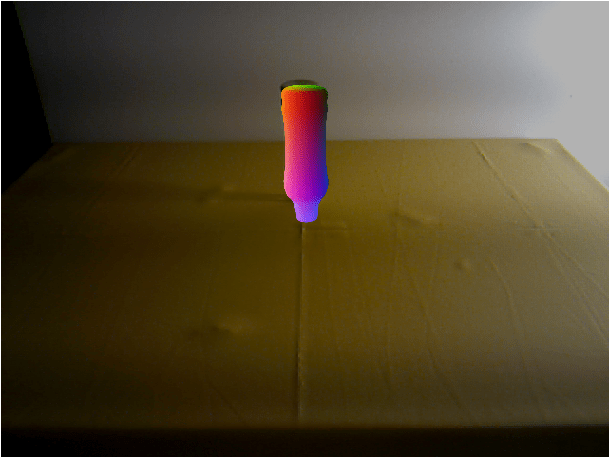}
    % \put(6,43){\small{\textcolor{magenta}{$\mathbf{Q}$}}}
    % \put(54,43){\small{\textcolor{magenta}{$\mathbf{A}$}}}
    \end{overpic}
\end{minipage}
\vspace*{0.2mm}
\begin{minipage}{0.17\textwidth}
    \begin{overpic}[width=1\textwidth]{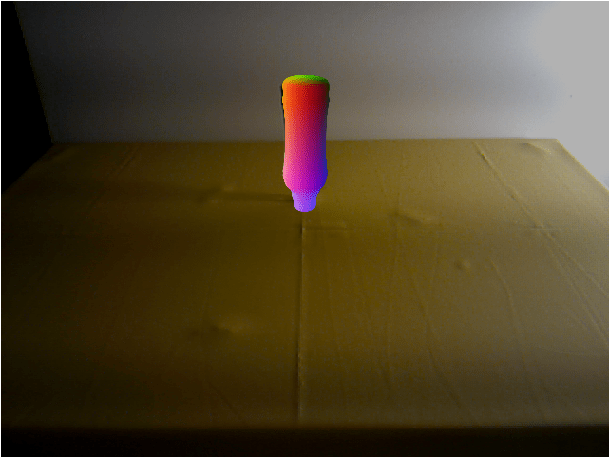}
    % \put(6,43){\small{\textcolor{magenta}{$\mathbf{Q}$}}}
    % \put(54,43){\small{\textcolor{magenta}{$\mathbf{A}$}}}
    \end{overpic}
\end{minipage}
\vspace*{1.5mm}
\begin{minipage}{\textwidth}
    \small \centering{(b) Prompt: \texttt{Green plastic bottle}}
\end{minipage}

\hspace*{0.00mm}
\begin{minipage}{0.17\textwidth}
    \begin{overpic}[width=1\textwidth]{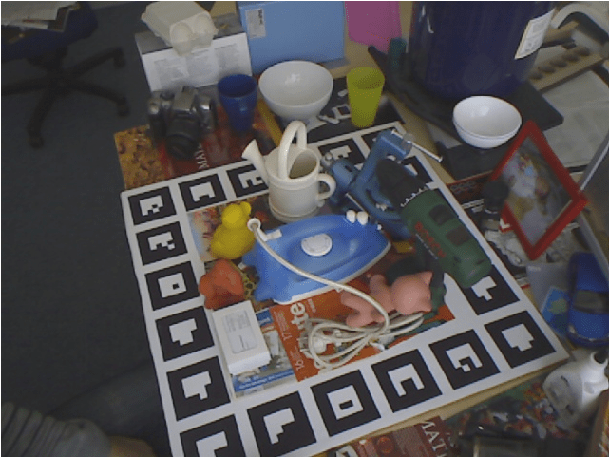}
    % \put(6,43){\small{\textcolor{magenta}{$\mathbf{Q}$}}}
    % \put(54,43){\small{\textcolor{magenta}{$\mathbf{A}$}}}
    \end{overpic}
\end{minipage}
\begin{minipage}{0.17\textwidth}
    \begin{overpic}[width=1\textwidth]{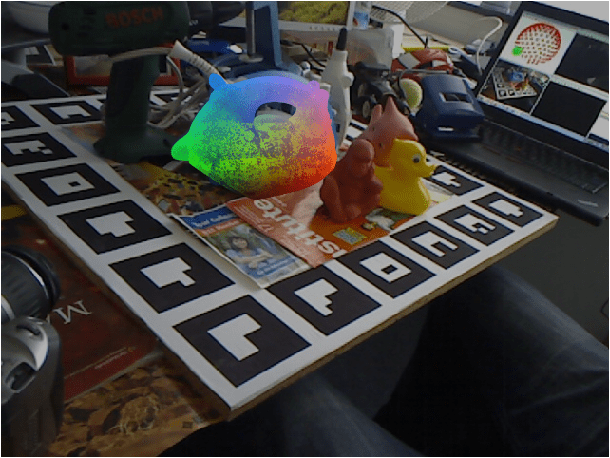}
    % \put(6,43){\small{\textcolor{magenta}{$\mathbf{Q}$}}}
    % \put(54,43){\small{\textcolor{magenta}{$\mathbf{A}$}}}
    \end{overpic}
\end{minipage}
\begin{minipage}{0.17\textwidth}
    \begin{overpic}[width=1\textwidth]{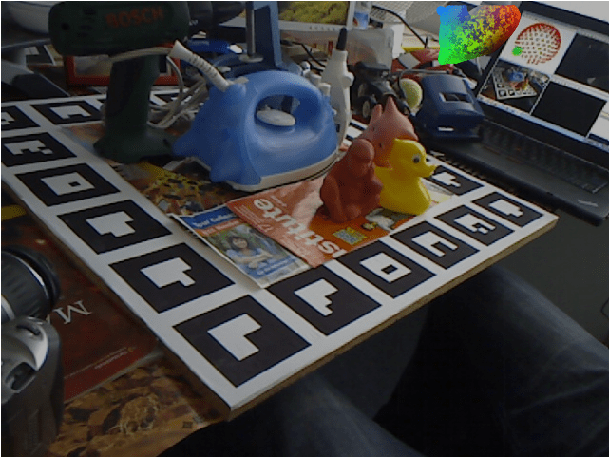}
    % \put(6,43){\small{\textcolor{magenta}{$\mathbf{Q}$}}}
    % \put(54,43){\small{\textcolor{magenta}{$\mathbf{A}$}}}
    \end{overpic}
\end{minipage}
\begin{minipage}{0.17\textwidth}
    \begin{overpic}[width=1\textwidth]{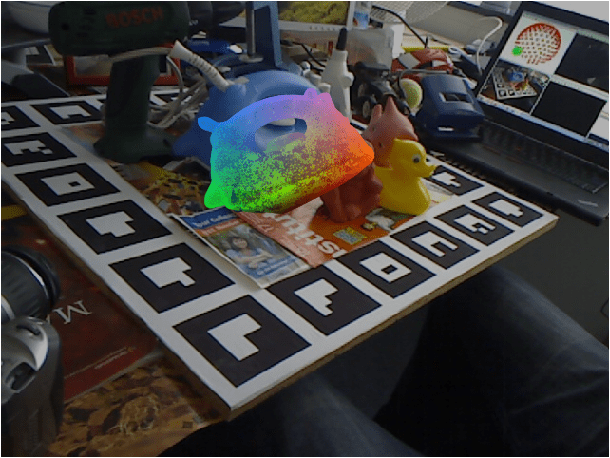}
    % \put(6,43){\small{\textcolor{magenta}{$\mathbf{Q}$}}}
    % \put(54,43){\small{\textcolor{magenta}{$\mathbf{A}$}}}
    \end{overpic}
\end{minipage}
\vspace*{0.2mm}
\begin{minipage}{0.17\textwidth}
    \begin{overpic}[width=1\textwidth]{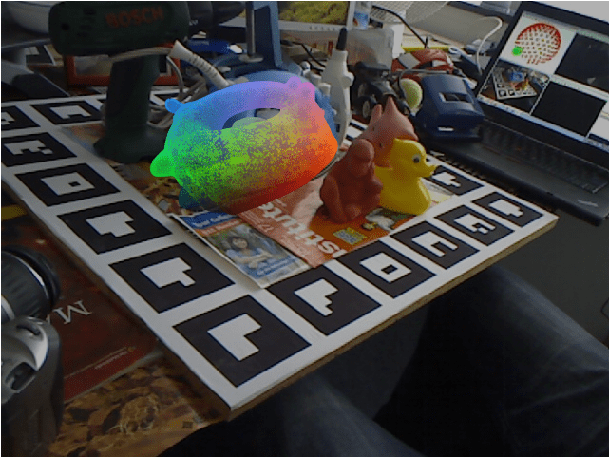}
    % \put(6,43){\small{\textcolor{magenta}{$\mathbf{Q}$}}}
    % \put(54,43){\small{\textcolor{magenta}{$\mathbf{A}$}}}
    \end{overpic}
\end{minipage}
\vspace*{1.5mm}
\begin{minipage}{\textwidth}
  \small  \centering{(c) Prompt: \texttt{Light blue iron}}
\end{minipage}

\hspace*{0.00mm}
\begin{minipage}{0.17\textwidth}
    \begin{overpic}[width=1\textwidth]{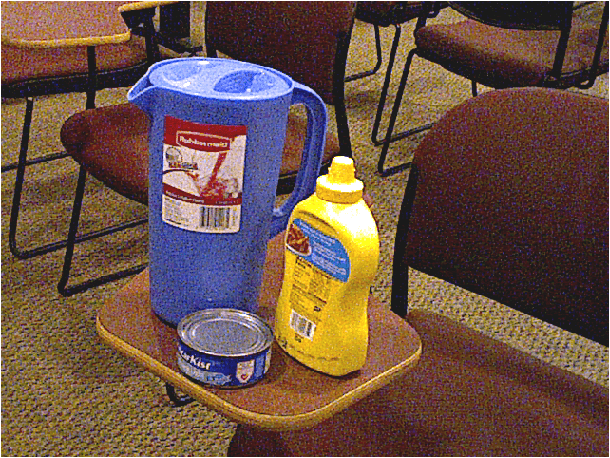}
    % \put(6,43){\small{\textcolor{magenta}{$\mathbf{Q}$}}}
    % \put(54,43){\small{\textcolor{magenta}{$\mathbf{A}$}}}
    \end{overpic}
\end{minipage}
\begin{minipage}{0.17\textwidth}
    \begin{overpic}[width=1\textwidth]{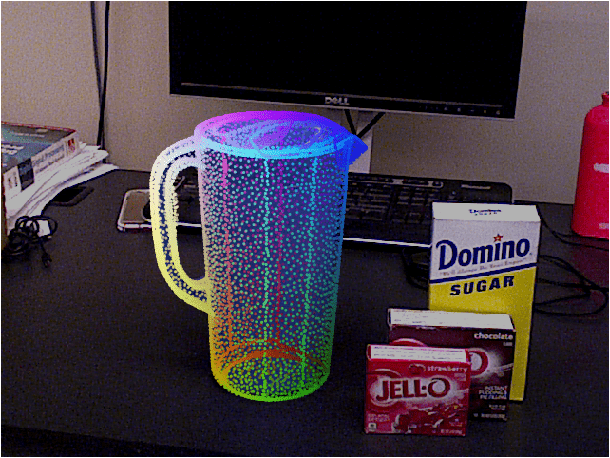}
    % \put(6,43){\small{\textcolor{magenta}{$\mathbf{Q}$}}}
    % \put(54,43){\small{\textcolor{magenta}{$\mathbf{A}$}}}
    \end{overpic}
\end{minipage}
\begin{minipage}{0.17\textwidth}
    \begin{overpic}[width=1\textwidth]{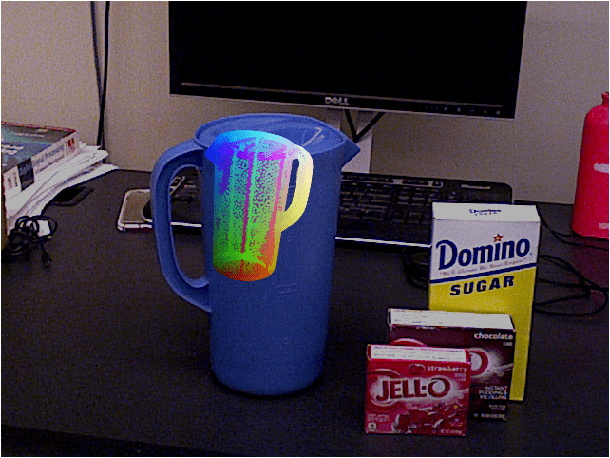}
    % \put(6,43){\small{\textcolor{magenta}{$\mathbf{Q}$}}}
    % \put(54,43){\small{\textcolor{magenta}{$\mathbf{A}$}}}
    \end{overpic}
\end{minipage}
\begin{minipage}{0.17\textwidth}
    \begin{overpic}[width=1\textwidth]{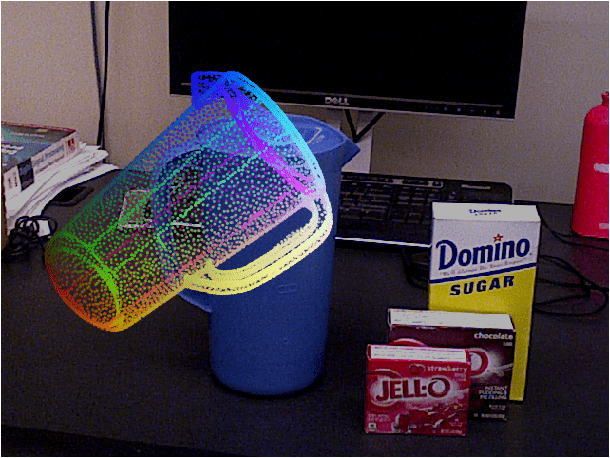}
    % \put(6,43){\small{\textcolor{magenta}{$\mathbf{Q}$}}}
    % \put(54,43){\small{\textcolor{magenta}{$\mathbf{A}$}}}
    \end{overpic}
\end{minipage}
\vspace*{0.2mm}
\begin{minipage}{0.17\textwidth}
    \begin{overpic}[width=1\textwidth]{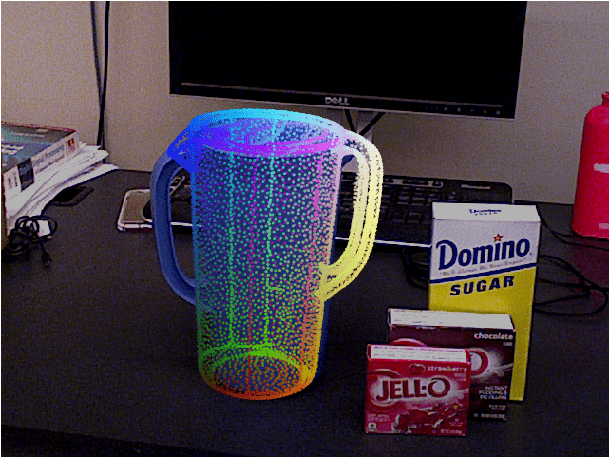}
    % \put(6,43){\small{\textcolor{magenta}{$\mathbf{Q}$}}}
    % \put(54,43){\small{\textcolor{magenta}{$\mathbf{A}$}}}
    \end{overpic}
\end{minipage}
\vspace*{1.5mm}
\begin{minipage}{\textwidth}
   \small \centering{(d) Prompt: \texttt{Blue plastic water jug}}
\end{minipage}

%% file: main/tables/ablation_oryon.tex
\newcolumntype{g}{>{\columncolor{mygray}}r}
\begin{table*}[t!]
    \centering
    \tabcolsep 3pt
    \caption{
    We show the results obtained by ablation the architectural components of Oryon, with the baseline at row 8 to better compare with the other results.
    All the results are computed with GroundingDino as detector prior and our predicted mask as segmentation prior.
    All metrics are the higher the better.
    }
    \vspace{-3mm}
    \label{tab:ablation_oryon}
    \resizebox{\textwidth}{!}{%
        \begin{tabular}{rccccccgggrrrgggrrrggg}

        \toprule
        & \multicolumn{6}{c}{Components} & \multicolumn{3}{c}{\cellcolor{mygray}REAL275} & \multicolumn{3}{c}{Toyota-Light} & \multicolumn{3}{c}{\cellcolor{mygray}Linemod} & \multicolumn{3}{c}{YCB-Video} & \multicolumn{3}{c}{\cellcolor{mygray}Average} \\
        & Crop & Loss Hyp. & Our \decoder & Fusion & Guidance & VLM & AR & ADD & mIoU & AR & ADD & mIoU & AR & ADD & mIoU & AR & ADD & mIoU & AR & ADD & mIoU \\
        \toprule
        \small{\color{gray} 1} & & & {\cmark} & Ours & {DINO} & {GDino} & 35.8 & 22.2 & 61.2 & 35.4 & 27.3 & 83.9 & 5.7 & 2.9 & 15.9 & 6.7 & 4.5 & 32.6  & 20.9 & 14.2 & 48.4 \\
        
        \small{\color{gray} 2} & {\cmark} & & {\cmark} & Ours & {DINO} & {GDino} & 43.5 & 30.8 & 80.8 & 28.6 & 15.9 & 82.3 & 15.6 & 14.2 & 51.4 & 13.6 & 8.8 & 67.6  & 25.3 & 17.4 & 70.5 \\
        \small{\color{gray} 3} & & {\cmark} & {\cmark} & Ours & {DINO} & {GDino} & 26.2 & 12.6 & 64.8 & 30.4 & 19.6 & 82.2 & 6.6 & 4.7 & 19.4 & 8.0 & 2.7 & 36.5 & 17.8 & 9.9 & 50.7 \\

        \small{\color{gray} 4} & {\cmark} & {\cmark} & {\cmark} & - & {DINO} & {GDino} & 51.2 & 39.0 & 74.8 & 31.2 & 19.2 & 77.9 & 19.9 & 18.2 & 48.0 & 13.7 & 5.7 & 55.6  & 29.0 & 20.5 & 64.1 \\
        \small{\color{gray} 5} & {\cmark} & {\cmark} & {\cmark} & Oryon & {DINO} & {GDino} & 49.2 & 41.8 & 79.2 & 32.9 & 24.4 & 80.9 & 21.5 & 19.8 & 51.3 & 18.9 & 9.4 & 60.4  & 30.6 & 23.8 & 67.9 \\
        \small{\color{gray} 6} & {\cmark} & {\cmark} & & Ours & {DINO} & {GDino} & 50.5 & 41.2 & 81.8 & 32.7 & 24.4 & 82.5 & 21.9 & 20.1 & 53.0 & 20.4 & 13.2 & 67.6  & 31.4 & 24.7 & 71.6 \\
        \small{\color{gray} 7} & {\cmark} & {\cmark} & {\cmark} & Ours & - & {GDino} & 48.2 & 32.6 & 80.3 & 32.3 & 23.1 & 82.3 & 21.5 & 20.1 & 52.2 & 20.6 & 12.8 & 67.6  & 30.6 & 22.2 & 70.6 \\
        \midrule
        \small{\color{gray} 8} & {\cmark} & {\cmark} & {\cmark} & Ours & {DINO} & {GDino} & 57.9 & 51.6 & 81.3 & 33.0 & 25.1 & 82.1 & 22.0 & 20.4 & 67.6 & 20.6 & 12.3 & 52.0 & 33.4 & 27.3 & 70.7 \\
        \midrule
        \small{\color{gray} 9} & {\cmark} & {\cmark} & {\cmark} & Ours & {Swin-B} & {GDino} & 49.1 & 46.2 & 81.3 & 33.2 & 24.2 & 82.3 & 21.9 & 20.3 & 51.9 & 21.0 & 13.8 & 68.5  & 31.3 & 26.1 & 71.0 \\
   
        \small{\color{gray} 10} & {\cmark} & {\cmark} & {\cmark} & Ours & {Swin-B} & {CLIP} & 37.3 & 19.7 & 72.0 & 26.9 & 13.4 & 69.1 & 16.7 & 14.2 & 53.8 & 11.4 & 4.9 & 54.5  & 23.1 & 13.1 & 62.3 \\
        \small{\color{gray} 11} & {\cmark} & {\cmark} & {\cmark} & Ours & {Swin-B} & {ALIGN} & 40.0 & 21.6 & 77.8 & 26.5 & 12.8 & 71.1 & 16.2 & 13.3 & 53.7 & 11.1 & 4.7 & 55.6  & 23.4 & 13.1 & 64.6 \\
        
        \bottomrule
    \end{tabular}
}

\end{table*}

%% file: main/tables/ablation_prompt.tex
\newcolumntype{g}{>{\columncolor{mygray}}r}
\begin{table*}[t!]
    \centering
    \tabcolsep 3pt
    \caption{
    We report the results obtained by changing the prompts at test time with an alternative version, which can have an incorrect description (Misleading), show only the object name (Generic), or show only the object description without the name (No name).
    Crop priors can be Oracle or predicted by GroundingDino~\cite{liu2023groundingdino}
    Segmentation priors can be Oracle or predicted by \acronym.
    All metrics are the higher the better.
    }
    \vspace{-3mm}
    \label{tab:abl_prompts}
    \resizebox{\textwidth}{!}{%
    \begin{tabular}{rlllgggrrrgggrrrggg}
        \toprule
        & \multirow{2}{*}{Prompt type} & \multirow{2}{*}{Crop} & \multirow{2}{*}{Mask} & \multicolumn{3}{c}{\cellcolor{mygray}REAL275} & \multicolumn{3}{c}{Toyota-Light} & \multicolumn{3}{c}{\cellcolor{mygray}Linemod} & \multicolumn{3}{c}{YCB-Video} & \multicolumn{3}{c}{\cellcolor{mygray}Average}  \\
        & & & & AR & ADD & mIoU & AR & ADD & mIoU & AR & ADD & mIoU & AR & ADD & mIoU & AR & ADD & mIoU \\
        \toprule
\small{\color{gray} \scriptsize 1}  &  \multirow{2}{*}{No Name}  & \oracle{Oracle} & \oracle{Oracle} & \oracle{55.0} & \oracle{46.2} & \oracle{100.0} & \oracle{36.8} & \oracle{28.9} & \oracle{100.0} & \oracle{34.4} & \oracle{28.1} & \oracle{100.0} & \oracle{28.2} & \oracle{22.7} & \oracle{100.0} & \oracle{38.6} & \oracle{31.5} & \oracle{100.0}\\
        \small{\color{gray} \scriptsize 2} & & GDino & Ours & 24.0 & 27.1 & 49.0 & 26.5 & 20.2 & 74.3 & 13.3 & 10.5 & 33.3 & 11.2 & 5.8 & 53.7  & 18.7 & 15.9 & 52.6 \\
        \midrule 
\small{\color{gray} \scriptsize 3}  &  \multirow{2}{*}{Misleading}  & \oracle{Oracle} & \oracle{Oracle} & \oracle{55.6} & \oracle{47.2} & \oracle{100.0} & \oracle{36.8} & \oracle{28.9} & \oracle{100.0} & \oracle{33.9} & \oracle{27.6} & \oracle{100.0} & \oracle{27.9} & \oracle{21.6} & \oracle{100.0} & \oracle{38.5} & \oracle{31.3} & \oracle{100.0}\\
        \small{\color{gray} \scriptsize 4} & & GDino & Ours & 40.1 & 38.2 & 75.9 & 28.5 & 19.6 & 76.7 & 7.3 & 4.0 & 20.2 & 13.9 & 7.0 & 50.6  & 22.5 & 17.2 & 55.8 \\
        \midrule
\small{\color{gray} \scriptsize 5}  &  \multirow{2}{*}{Generic}  & \oracle{Oracle} & \oracle{Oracle} & \oracle{55.6} & \oracle{47.5} & \oracle{100.0} & \oracle{36.8} & \oracle{28.1} & \oracle{100.0} & \oracle{33.7} & \oracle{27.6} & \oracle{100.0} & \oracle{28.0} & \oracle{21.9} & \oracle{100.0} & \oracle{38.5} & \oracle{31.3} & \oracle{100.0}\\
        \small{\color{gray} \scriptsize 6} & & GDino & Ours & 40.3 & 37.6 & 75.6 & 36.1 & 27.7 & 85.3 & 13.6 & 10.9 & 33.1 & 14.3 & 4.3 & 54.2  & 26.1 & 20.1 & 62.0 \\
        \midrule
\small{\color{gray} \scriptsize 7}  &  \multirow{2}{*}{Standard}  & \oracle{Oracle} & \oracle{Oracle} & \oracle{57.7} & \oracle{49.8} & \oracle{100.0} & \oracle{37.4} & \oracle{28.4} & \oracle{100.0} & \oracle{34.4} & \oracle{27.6} & \oracle{100.0} & \oracle{28.6} & \oracle{22.6} & \oracle{100.0} & \oracle{39.5} & \oracle{32.1} & \oracle{100.0}\\
        \small{\color{gray} \scriptsize 8} & & GDino & Ours & 57.9 & 51.6 & 81.3 & 33.0 & 25.1 & 82.1 & 22.0 & 20.4 & 67.6 & 20.6 & 12.3 & 52.0 & 33.4 & 27.3 & 70.7 \\
        \bottomrule        
    \end{tabular}  
 }
\end{table*}

%% file: main/sections/5_conclusion.tex
\section{Conclusions}\label{sec:conclusion}

We presented \acronym, an approach that significantly improves upon previous open-vocabulary 6D pose estimation methods, by increasing the feature map resolution and providing more accurate matches to perform registration. 
Our experiments show that \acronym obtains good performances also in scenarios with unusual objects (Linemod) and mild occlusions (YCB-Video).
The ablation studies show that the token-wise representation provided by the updated VLM, together with the new fusion strategy, greatly benefit \acronym. 
The improvement is consistent in terms of generalisation capabilities and robustness to prompt noise as well.

\acronym's limitations are the need for depth maps and intrinsic camera parameters to perform registration.
Such data requirements could be relaxed by exploring monocular depth estimation methods such as DepthAnything~\cite{depthanything} on each RGB image.
While \acronym is more robust to suboptimal prompts than Oryon, the resulting drop in performance is still significant. 
Moreover, the variety of prompt usable at test time is limited by the training data, which provides prompts without descriptions.
To enrich the prompts at training time, LLMs or image captioners could be used on image samples to provide prompts that also include a description of the object’s colour and physical attributes.

\noindent\textbf{Acknowledgements.} This work was supported by the European Union’s Horizon Europe research and innovation programme under grant agreement No 101058589 (AI-PRISM), and it made use of time on the Tier 2 HPC facility JADE2, funded by EPSRC (EP/T022205/1).

%% file: main/sections/6_bios.tex
\vspace{-0.5cm}
\begin{IEEEbiography}[{\includegraphics[width=1in,height=1.25in,clip,keepaspectratio]{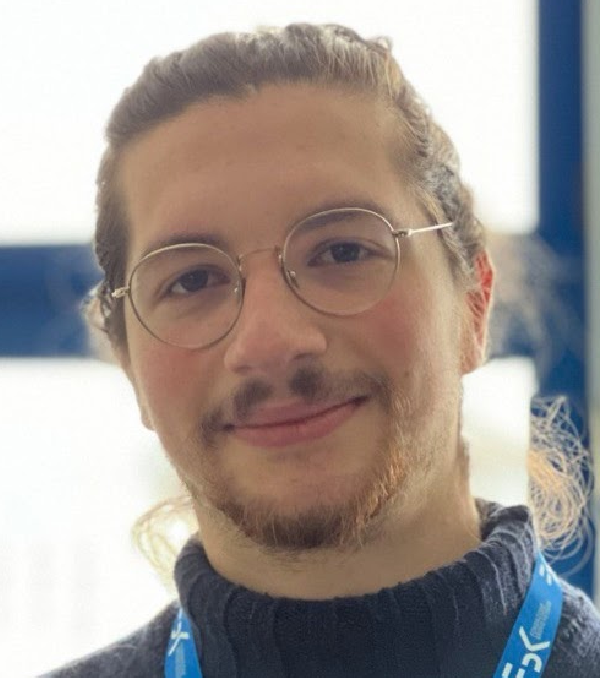}}]{Jaime Corsetti}
is a PhD student at University of Trento, affiliated with the Technologies of Vision Lab in Fondazione Bruno Kessler in Trento, Italy.  He received his MsC Degree in Artificial Intelligence Systems at the University of Trento in October 2023. His research interests are in object 6D pose estimation and 3D scene understanding.
\end{IEEEbiography}
\vspace{-0.5cm}
\begin{IEEEbiography}[{\includegraphics[width=1in,height=1.25in,clip,keepaspectratio]{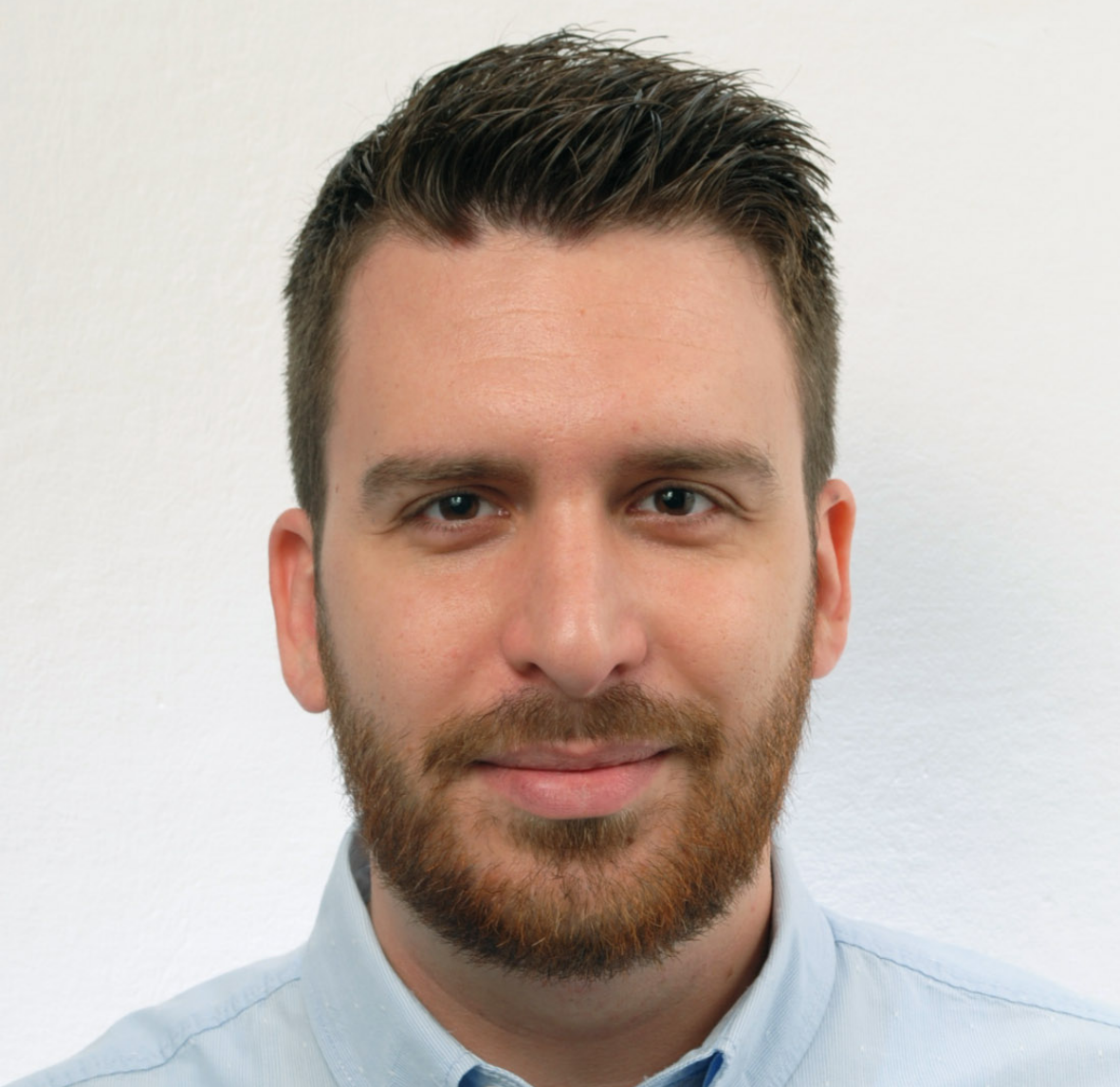}}]{Davide Boscaini}
is a Research Scientist at the Technologies of Vision Lab of the Fondazione Bruno Kessler in Trento, Italy. He received his PhD in Computational Science from the Universit\`{a} della Svizzera italiana in Lugano, Switzerland, under the supervision of Michael Bronstein. He has been a pioneer in the field of geometric deep learning. His current research interests include 3D perception and understanding, with a focus on object 6D pose estimation and 3D scene understanding.
\end{IEEEbiography}
\vspace{-0.5cm}
\begin{IEEEbiography}[{\includegraphics[width=1in,height=1.25in,clip,keepaspectratio]{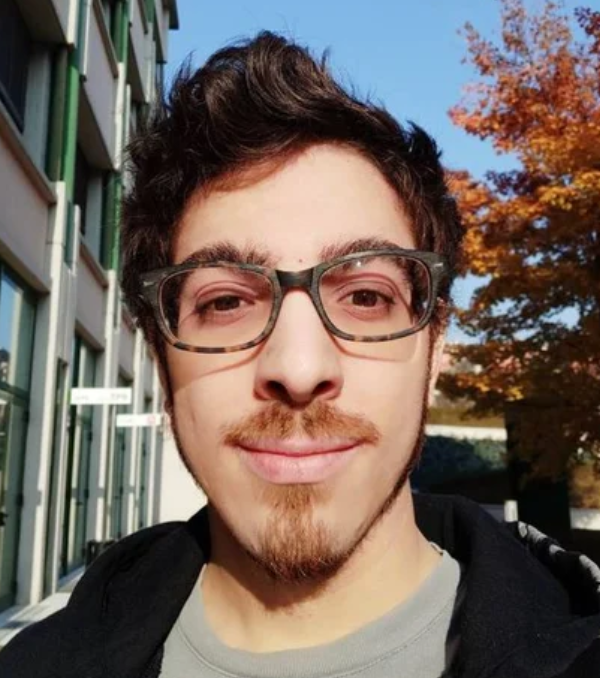}}]{Francesco Giuliari} 
is a Researcher at the Technologies of Vision Centre in Fondazione Bruno Kessler in Trento, Italy.
He received his PhD degree in Deep Learning from the University of Genoa in 2024. His research interests include visual scene understanding using scene graphs, and vision-based robot navigation using classical and deep learning planners. 
\end{IEEEbiography}
\vspace{-0.5cm}
\begin{IEEEbiography}[{\includegraphics[width=1in,height=1.25in,clip,keepaspectratio]{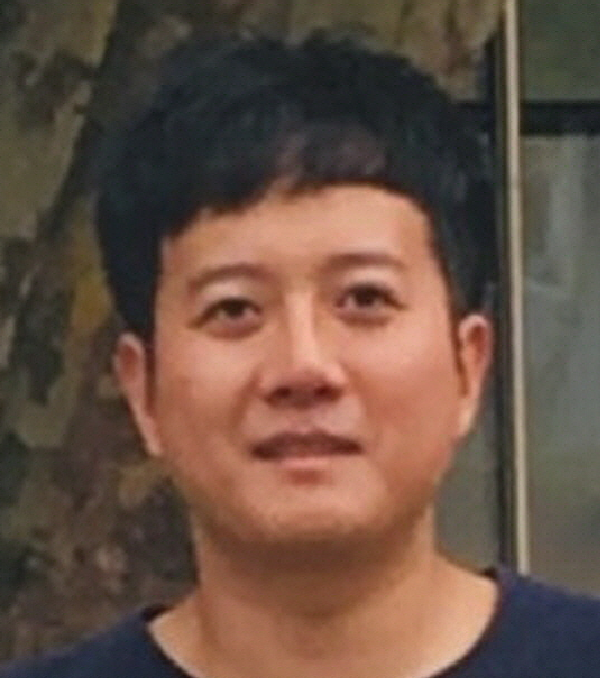}}]{Changjae Oh} 
is a Lecturer at the School of Electronic Engineering and Computer Science and the Centre for Intelligent Sensing from Queen Mary University of London, UK. He received his PhD in Electrical and Electronic Engineering from Yonsei University, Seoul, South Korea, in 2018. 
From 2018 to 2019, he was a Postdoctoral Research Assistant at Queen Mary University of London, UK.
His current research interests include embodied agents and vision-based robot manipulation.
\end{IEEEbiography}
\vspace{-0.5cm}
\begin{IEEEbiography}[{\includegraphics[width=1in,height=1.25in,clip,keepaspectratio]{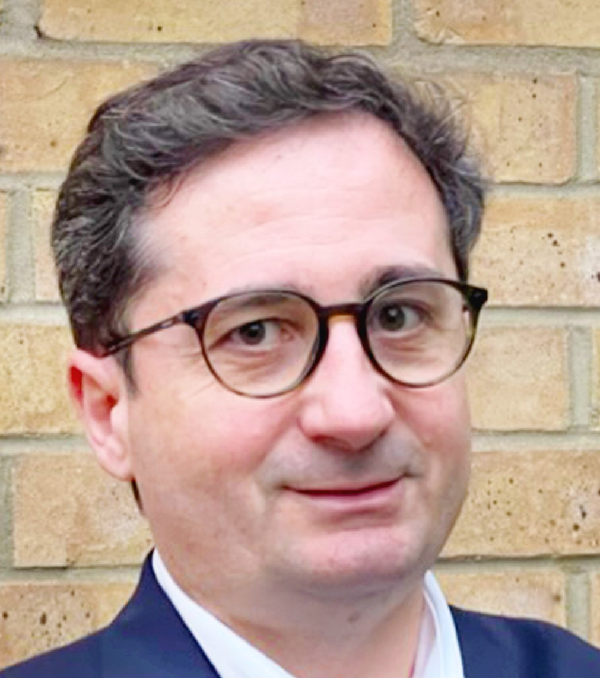}}]{Andrea Cavallaro} 
is the director of the Idiap Research Institute  and a Full Professor at École Polytechnique Fédérale de Lausanne (EPFL), Switzerland. He has been a Full Professor at Queen Mary University of London since 2010. He is a Fellow of the International Association for Pattern Recognition. His research interests include machine learning for multimodal perception, computer vision, audio processing and information privacy.
\end{IEEEbiography}
\vspace{-0.5cm}
\begin{IEEEbiography}[{\includegraphics[width=1in,height=1.25in,clip,keepaspectratio]{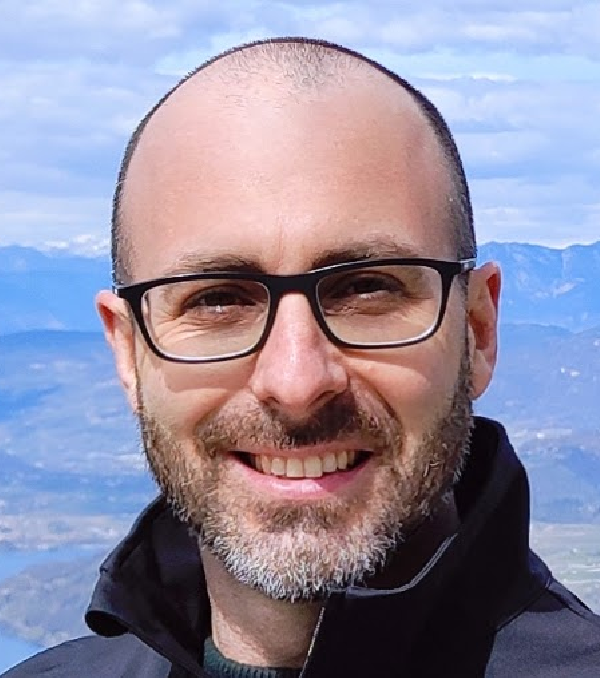}}]{Fabio Poiesi} is the Head of the Technologies of Vision (TeV) Lab, Fondazione Bruno Kessler, Trento, Italy.
He received the PhD degree from the Queen Mary University of London, U.K. and was a post-doctoral researcher with the Queen Mary University of London before moving to TeV. His research interests include 3D scene understanding, object detection and tracking, and extended reality. 
\end{IEEEbiography}